\documentclass[11pt,table]{article}
\pdfoutput=1

\usepackage{amsthm}

\usepackage{titling}
\usepackage[utf8]{inputenc} 
\usepackage[T1]{fontenc}    
\usepackage[in]{fullpage}
\usepackage{microtype}
\usepackage{graphicx}
\usepackage{amsmath}
\usepackage{amssymb}
\usepackage{amsthm}
\usepackage{enumitem}
\usepackage{xfrac}
\usepackage{parskip}
\usepackage{numprint}
\usepackage{booktabs} 
\usepackage[svgnames,dvipsnames,color,table]{xcolor}
\usepackage[textsize=scriptsize,textwidth=1.9cm]{todonotes}
\usepackage{xspace}
\usepackage{float}
\usepackage{subcaption}
\usepackage{etoolbox}
\usepackage{comment}
\usepackage{etoc}
\usepackage{wrapfig}
\usepackage{placeins}
\usepackage{makecell}

\newtoggle{isarxiv}
\toggletrue{isarxiv}

\definecolor{DarkRed}{rgb}{0.5,0.1,0.1}
\definecolor{DarkBlue}{rgb}{0.1,0.1,0.5}
\usepackage{hyperref}
\hypersetup{
		colorlinks=true,
		linkcolor=blue!70!black,
		citecolor=blue!70!black,
		urlcolor=blue!70!black}

\usepackage[numbers,sort]{natbib}
\usepackage{adjustbox}
\usepackage{multirow}
\usepackage{color, colortbl}
\definecolor{Gray}{gray}{0.9}
\usepackage{fnpct}

\begin{document}
\date{}
\title{On the Connection between Pre-training Data Diversity and Fine-tuning Robustness}
\author{
    Vivek Ramanujan\thanks{Equal Contribution}\textsuperscript{ }\thanks{University of Washington}
    \hspace{1cm}
    Thao Nguyen\footnotemark[1]\textsuperscript{ }\footnotemark[2]
    \and
    Sewoong Oh\footnotemark[2]\textsuperscript{ }\thanks{Google Research}
    \hspace{1cm}
    Ludwig Schmidt\footnotemark[2]\textsuperscript{ }\thanks{Allen Institute for Artificial Intelligence}\textsuperscript{ }\thanks{LAION}
    \hspace{1cm}
    Ali Farhadi\footnotemark[2] \\
}
\maketitle
\vspace{-.2cm}

\begin{abstract}
Pre-training has been widely adopted in deep learning to improve model performance, especially when the training data for a target task is limited. In our work, we seek to understand the implications of this training strategy on the generalization properties of downstream models. 
More specifically, we ask the following question: how do properties of the pre-training distribution affect the robustness of a fine-tuned model? The properties we explore include the label space, label semantics, image diversity, data domains, and data quantity of the pre-training distribution.
We find that the primary factor influencing downstream effective robustness~\cite{taori2020measuring} is data quantity, while other factors have limited significance. For example, reducing the number of ImageNet pre-training classes by 4$\times$ while increasing the number of images per class by 4$\times$ (that is, keeping total data quantity fixed) does not impact the robustness of fine-tuned models. We demonstrate our findings on pre-training distributions drawn from various natural and synthetic data sources, primarily using the iWildCam-WILDS distribution shift as a test for downstream robustness.
\end{abstract}

\section{Introduction}
\label{sec:intro}
Transfer learning is a popular technique to deal with data scarcity, improve training speed, or transfer useful inductive biases that can benefit downstream tasks \cite{mormont2018comparison, conneau2018senteval, de2018clinically}. In the domain of computer vision, pre-training on ImageNet in particular has been the de-facto standard for obtaining features to solve a wide range of vision tasks, such as object detection \cite{ren2015faster, dai2016r, huang2017speed}, segmentation \cite{chen2017deeplab, he2017mask}, and action recognition \cite{simonyan2014two}. 
While there exists previous work that seeks to pinpoint specific properties of ImageNet-trained features that benefit downstream performance \cite{huh2016makes, kornblith2019better, kolesnikov2020big, salman2020adversarially}, the analysis is often done with respect to model accuracy. Our work instead examines the robustness of fine-tuned models to natural distribution shifts. Instead of looking at architecture variations and pre-training algorithms as done in prior work~\cite{salman2020adversarially, guo2019spottune, you2020co}, we focus on the role of the pre-training data. This data-centric approach has been validated by past work~\cite{miller2021accuracy, nguyen2022quality, fang2022data}, which show that the training data distribution plays a larger role than training method or architecture in influencing model robustness.

Robustness under distribution shifts is a fundamental concern for producing reliable machine learning systems: a model can perform in unexpected and undesirable ways when there is a mismatch between the data distribution encountered in deployment and the one on which the model is trained~\cite{rosenfeld2018elephant, koh2021wilds}. For example, a self-driving car should be able to generalize to a wide variety of weather scenarios to be considered safe, some of which it may not have seen during training. In our work, we focus on these forms of \textit{natural} distribution shifts \cite{recht2019imagenet, koh2021wilds}---named so because they are induced by real-world processes---and study what aspects of the source dataset could help fine-tuned models become more robust to these shifts. We tackle this question along five different ablation axes: \textbf{(i)} Data quantity, \textbf{(ii)} Label granularity, \textbf{(iii)} Label semantics, \textbf{(iv)} Image diversity, and \textbf{(v)} Data sources. Most of our experiments revolve around the supervised learning setting, which makes controlled experiments on the pre-training distribution more tractable and systematic. Through a better understanding of the interplay between various properties of the pre-training distribution and downstream robustness, we seek to establish guidelines for constructing better pre-training datasets for fine-tuning. 

Previous work by \citet{miller2021accuracy} experimented with a wide range of natural distribution shifts and found that pre-training on ImageNet yields the biggest improvement in robustness for the task of iWildCam-WILDS classification~\cite{koh2021wilds, beery2020iwildcam}. Consequently, we use iWildCam-WILDS as a probe to evaluate how our interventions with the ImageNet pre-training distribution would alter the robustness trend uncovered in this previous work. We also analyze the use of other pre-training data sources that may differ significantly from ImageNet in both semantic content and data collection methodology.
\begin{figure}
    \centering
    \includegraphics[width=\textwidth]{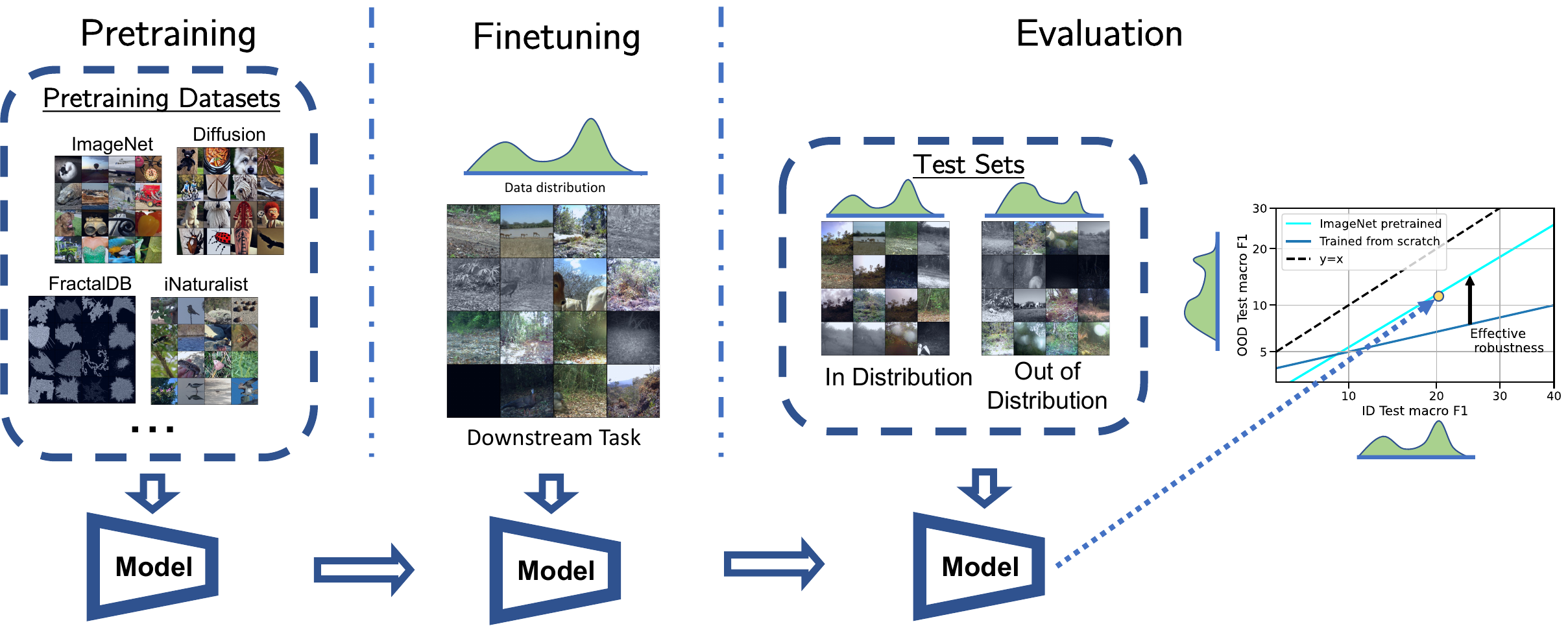}
    \caption{A summary of our experimental pipeline. We pre-train a model on a variety of different data distributions and evaluate its effective robustness after fine-tuning on a downstream task (i.e., iWildCam). By examining many models in this manner, we can determine empirical properties of the pre-training distribution that are important for fine-tuning robustness.}
    \label{fig:teaser}
\end{figure}
Our main findings can be summarized as follows:
\begin{enumerate}[label=(\roman*), topsep=0.5pt,align=left, leftmargin=0pt, labelindent=\parindent, listparindent=\parindent, itemindent=!]
    \item \textbf{Data quantity.} Pre-training with more data helps boost robustness. However, we do not need a lot of pre-training data to see significant robustness gains: using 25K images subsampled from either ImageNet or iNaturalist, which is 6$\times$ smaller than the size of the fine-tuning dataset, already offers noticable robustness improvements.  
    \item \textbf{Label granularity.} Making pre-training labels more coarse-grained lowers transfer robustness. The effect is less significant than altering data quantity: extreme reduction in label granularity (e.g., using 5 coarse classes instead of 1000 fine-grained classes) still preserves some of the robustness gains compared to training from scratch. 
    \item \textbf{Label semantics.} Given enough data and labels, pre-training on more semantically similar classes does not have a notable impact on the robustness of fine-tuned models. In particular, we find that pre-training on the 600 inanimate object categories in ImageNet yields the same effective robustness as pre-training on the 400 animal categories, despite the fact that the downstream task consists of only animal categories.
    \item \textbf{Image diversity.} Given the same pre-training label set and data quantity, increasing per-class diversity (e.g., by including more subclasses) has no effect on transfer robustness. In addition, the trade-off between having more classes and more images per class is negligible if the total number of samples is kept constant. 
    \item \textbf{Data sources.} We find that natural data sources (i.e., ImageNet, iNaturalist) yield similar downstream robustness when controlling for data quantity. Pre-training with synthetic fractal data is less effective at the same data quantity regime but still has some robustness gain to offer compared to training from scratch. Synthetic natural-looking data (e.g., generated by Stable Diffusion~\cite{rombach2021highresolution}) can help close this gap between using natural data and synthetic fractal data.
\end{enumerate}

Overall we find that increasing pre-training data quantity and label granularity makes fine-tuned models more robust to distribution shifts. However, not all additional data is equally helpful. For instance, in the context of iWildCam-WILDS task, pre-training with natural-looking data offers much more robustness than using 10$\times$ more synthetic fractal data.
\section{Background}
The main motivation for our paper comes from the work by \citet{huh2016makes}, which investigates various factors in ImageNet training that affect the quality of the features used subsequently for transfer learning. For our investigation, we shift the focus from accuracy to robustness against distribution shift, which has been a long-standing issue in machine learning~\cite{quinonero2008dataset, szegedy2013intriguing, biggio2018wild, biggio2013evasion}.
In particular, we analyze the robustness of pre-trained features to natural distribution shifts observed in the real world through the iWildCam-WILDS benchmark~\cite{koh2021wilds}. 
Furthermore, in contrast to \citet{huh2016makes}, we experiment with a greater variety of more recent neural network architectures, in addition to exploring the use of synthetic pre-training data. 

\begin{figure}[t]
\begin{minipage}{0.48\textwidth}
\includegraphics[trim=0 0 0 0,clip,width=\linewidth]{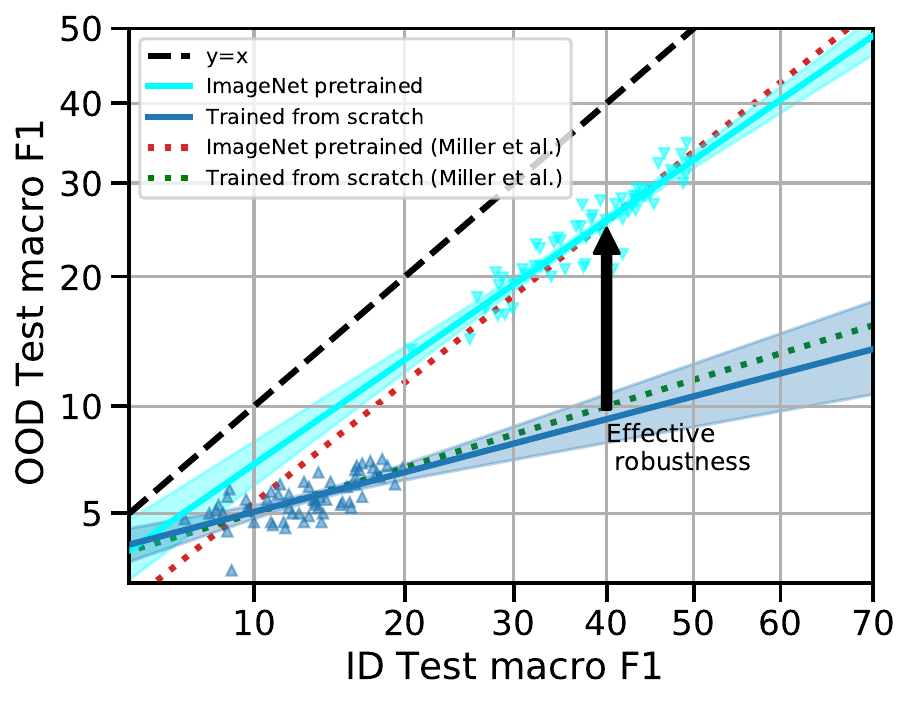}
\end{minipage}\hfill
\begin{minipage}{0.5\textwidth}
\caption{Effective robustness is defined as movement towards a classifier which is robust to distribution shift (i.e., line $y=x$). Using this metric, \citet{miller2021accuracy} observes that for the iWildCam-WILDS task, models pre-trained on ImageNet are much more robust than models trained from scratch. We reproduce these two trends and use them as points of reference for our subsequent experiments, in which we modify the pre-training distribution and observe how our interventions alter the robustness trend lines.
}
\label{fig:effective_robustness}
\end{minipage}
\vskip -1em
\end{figure}

A key goal in robustness is to reduce the impact of distribution shifts on the performance of a model.
If model performances on in- and out-of-distribution test sets are plotted along the $x$ and $y$-axes of a scatter plot respectively, then a more robust model would lie closer to the diagonal $y=x$ line. This notion of robustness was captured by \citet{taori2020measuring} under the term \emph{effective robustness}, which measures the difference between a model’s actual OOD performance and what could be predicted from its ID performance (Figure \ref{fig:effective_robustness}). 
\citet{miller2021accuracy} adopted this effective robustness framework and evaluated hundreds of models on various distribution shift settings. The authors observed that ID performance is highly correlated with OOD performance. This linear trend mapping ID to OOD performance, and how close it is to the $y=x$ line, is what we use in our work to compare the quality of the pre-trained features. 

More notably, \citet{miller2021accuracy} discovered that on the iWildCam dataset, models trained from scratch and those that have been pre-trained on ImageNet lie on distinct linear trends, with the latter exhibiting much more robustness.
We replicate these reported trends in Figure~\ref{fig:effective_robustness}. Motivated by this result, our work seeks to better understand what aspects of ImageNet pre-training contribute to the improved robustness on iWildCam, and how these aspects translate to other pre-training data sources.
\begin{figure}[]
\begin{minipage}{0.53\textwidth}
\includegraphics[trim=0 0 0 0,clip,width=\linewidth]{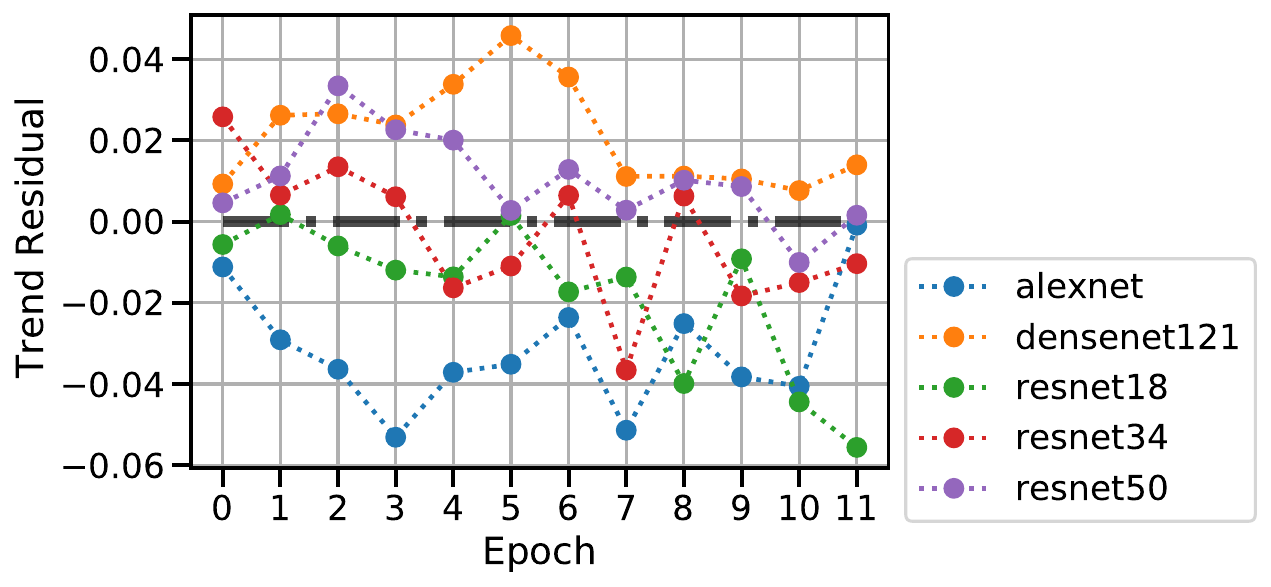}
\end{minipage}\hfill
\begin{minipage}{0.45\textwidth}
\caption{We visualize the residuals of various architectures after fitting a linear trend that predicts OOD accuracy from ID accuracy. All models are pre-trained on the full ImageNet dataset and fine-tuned on iWildCam for 12 epochs. We observe that overall the residuals fluctuate around the $y=0$ line and vary throughout the course of fine-tuning for most architectures.}
\label{fig:sanity_check}
\end{minipage}
\vskip -1.5em
\end{figure}

Previous work by \citet{andreassen2021evolution} has looked at effective robustness over the course of fine-tuning and found that pre-trained models exhibit high effective robustness in the middle of fine-tuning, which eventually decreases as the training proceeds. The paper also experimented with ImageNet as one of the pre-training data sources. In our investigation, as a sanity check to remove number of training epochs as a potential source of bias for the linear fit, we adopt the linear trend of models pre-trained on ImageNet and fine-tuned on iWildCam computed previously by \citet{miller2021accuracy} as the baseline. We then report the residuals from comparing actual OOD performance at different epochs to what could be predicted from the corresponding ID performance using this baseline. Refer to Figure \ref{fig:sanity_check} for more details. We find that in the context of iWildCam fine-tuning, at each epoch, the residuals from our architectures of choice concentrate around the $y=0$ line and exhibit no particular trend. This in turn allows us to vary the number of fine-tuning epochs as a hyperparameter, and obtain models covering a wide range of test performances for the scatter plots.

\section{Experimental Setup} \label{sec:setup}
As mentioned earlier, the downstream task of interest is wildlife classification with the iWildCam-WILDS dataset \cite{koh2021wilds}:
the input is a photo taken by a camera trap, and the output is one of 182 different animal species.
There are two test sets for evaluation: ID test data consists of images taken by the same camera traps as the training set, but on different days from the training and validation (ID) images.
In contrast, OOD test data contains images taken by a disjoint set of camera traps from training and validation (ID) images.
We include some examples of the geodiversity represented in each test split in Appendix Figure~\ref{fig:iwildcam_vals}.
Following \cite{koh2021wilds}, we report the macro F1 scores of the trained networks because this metric emphasizes performance on rare species, which is critical to the biodiversity monitoring application that the dataset was designed for.

\begin{table*}[t]
\centering
\resizebox{\textwidth}{!}{\begin{tabular}{ cccccc } 
  \hline
   & Training set size & Number of classes & Class distribution & Class hierarchy & Expert-labeled \\
  \hline
  ImageNet & 1,281,167 & 1,000 & Class-balanced & WordNet & No \\ 
  \hline
  iNaturalist & 579,184 & 5,089 & Long-tailed & Tree of life & Yes\\
  \hline
\end{tabular}\label{tab:inat_imagenet}}
\vskip -.25em
\caption{Differences between the ImageNet and iNaturalist datasets.}
\label{tab:inat_imagenet}
\end{table*}

\textbf{Pre-training datasets.} We use ImageNet~\cite{deng2009imagenet} and iNaturalist~\cite{van2018inaturalist} as the primary pre-training distributions of interest, given their hierarchical structures, complexity, and relevance to the downstream task. The two data sources also differ in many ways (Table~\ref{tab:inat_imagenet}), hence their pre-trained features make for an informative comparison. We also include experiments with synthetic pre-training data by using Stable Diffusion~\cite{rombach2021highresolution} and the FractalDB-1k dataset~\cite{KataokaIJCV2022}. We will elaborate on this in Section~\ref{subsection:inat_pre-training}.

\textbf{Network architectures.} To obtain data for plotting linear trends, we train a range of standard neural network architectures including ResNet \cite{he2016deep}, ResNext \cite{xie2017aggregated}, DenseNet \cite{iandola2014densenet}, AlexNet \cite{krizhevsky2012imagenet} and MobileNet-V3 \cite{howard2019searching}. In our scatter plots, besides varying the architectures, we also vary the number of fine-tuning epochs to obtain models with varying F1 scores. Appendix \ref{app:train_details} contains further training details. While our focus is on supervised pre-training, we also report some additional results with CLIP~\cite{radford2021learning} architecture in Section \ref{sec:self_supervised}.

In the subsequent sections, we detail different interventions made to the pre-training distribution to disentangle key properties of interest. We show the resulting linear trends in relation to the trends replicated from previous work \cite{miller2021accuracy}, which include models trained from scratch on iWildCam (solid blue line) as well as models pre-trained on ImageNet (solid cyan line). For each trend line, we show 95\% bootstrap confidence intervals for the linear fit.
\section{Experiment Results}
\subsection{Effect of Data Quantity}
\label{sec:data_quantity}
\begin{figure}
\centering
{\includegraphics[width=0.4\textwidth]{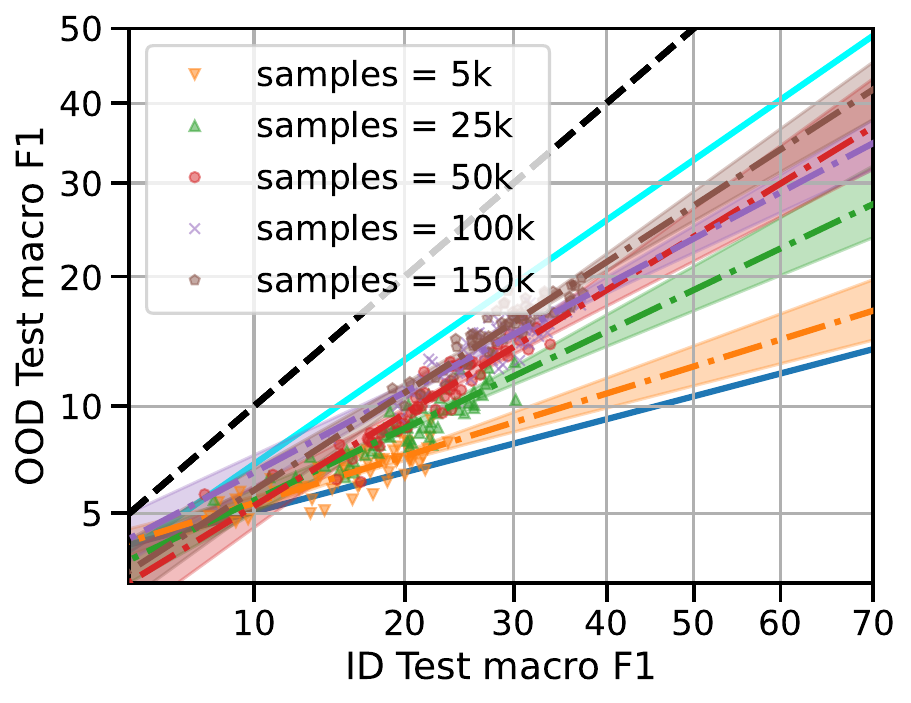}}
{\includegraphics[width=0.4\textwidth]{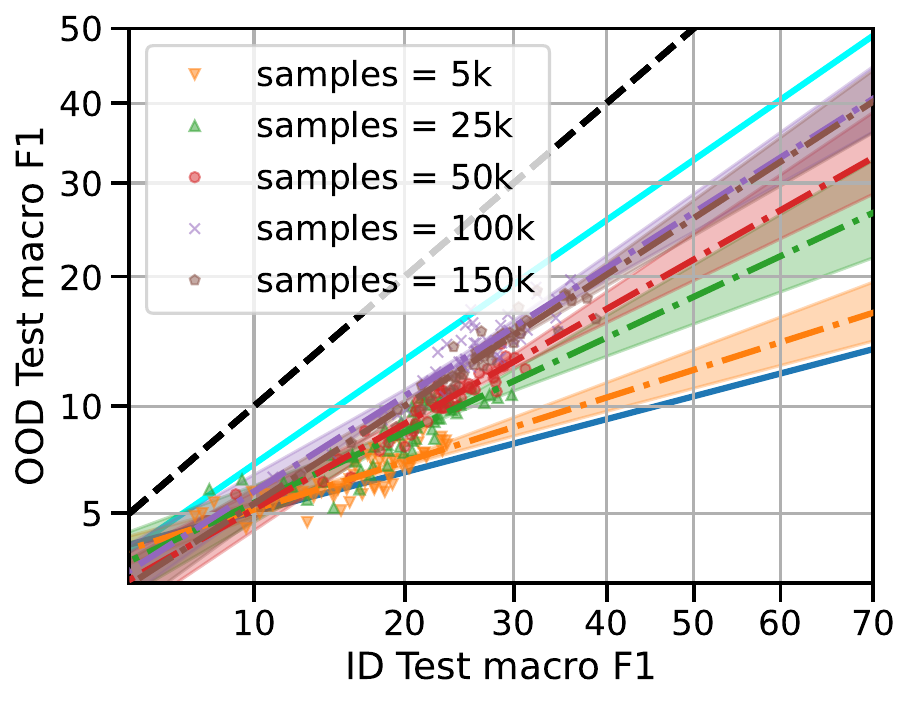}}
\vskip -.75em
\caption{Reducing the number of pre-training images randomly sampled from \textbf{(left)} ImageNet and \textbf{(right)} iNaturalist lowers the robustness linear trends of the fine-tuned models. However, using only 25K pre-training samples (green line) still yields significant robustness improvements compared to training from scratch on 129K iWildCam images (dark blue line).
We subsample iNaturalist and ensure class balance, only including 1000 classes with the most number of samples.}
\label{fig:subsample}
\end{figure}
First, we experiment with reducing the pre-training set size. To remove potential confounding effects from a long-tailed data distribution, we ensure that the class distribution of our pre-training datasets is uniform. ImageNet is already class-balanced, but this is not the case for iNaturalist \cite{van2018inaturalist}. We experiment with a 1000-class subset of iNaturalist using its most frequent classes. We further select images within each class uniformly at random so that the number of samples is the same across all classes. This results in a class-balanced training set of size 150K from iNaturalist. We repeat the same procedure to obtain subsets of size 100K, 50K, 25K and 5K. A similar subsampling process is done on ImageNet, using the full 1000-class dataset which already has a uniform label distribution.

In Figure \ref{fig:subsample}, we observe that reducing the data quantity during pre-training lowers the effective robustness of fine-tuned models. However, it is worth noting at 25K images, pre-training with subsampled ImageNet and iNaturalist data still produces much more robust models compared to training from scratch. This is 6$\times$ less data compared to what is used for fine-tuning.
As a sanity check, we find that using only 5K samples (i.e., 5 examples per class) during pre-training yields roughly the same level of robustness as training from scratch on iWildCam.

\subsection{Effect of Label Granularity} \label{sec:label_granularity}
\begin{figure}
\centering
{\includegraphics[width=0.4\textwidth]{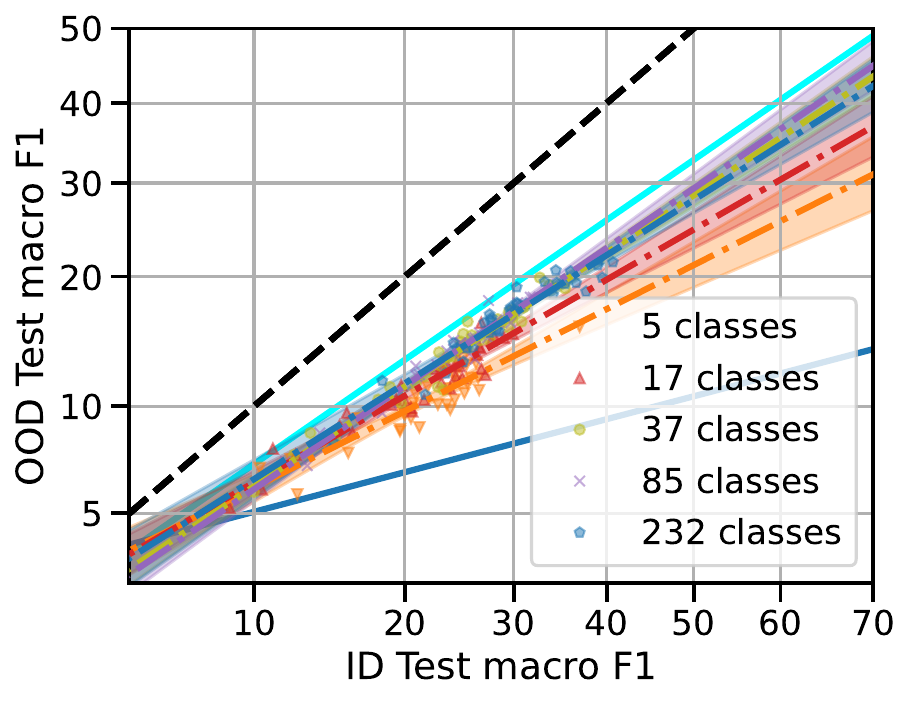}}
{\includegraphics[width=0.4\textwidth]{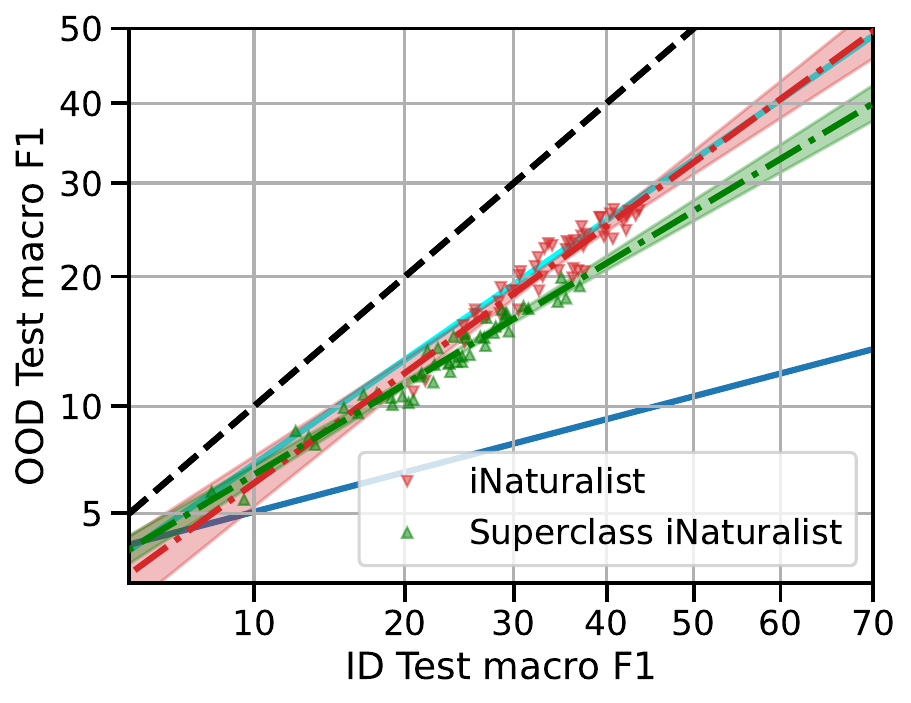}}
\vskip -.75em
\caption{Results of changing the label granularity of the pre-training task by combining classes according to some semantics hierarchy to form supersets, for \textbf{(left)} ImageNet and \textbf{(right)} iNaturalist. In general, this intervention lowers model robustness on downstream task. However, extreme reduction of the pre-training label space, e.g. by 200$\times$ in the case of ImageNet, still offers robustness gains compared to training from scratch.
}
\label{fig:max_depth}
\end{figure}

Next, we adapt a question raised previously by \citet{huh2016makes} to our investigation: how does varying the number of pre-training classes affect downstream robustness?
Following \cite{huh2016makes}, we construct \emph{supersets} of classes in ImageNet using the WordNet hierarchy.
We use the maximum of the shortest path distance from the root of WordNet to a label to compute the depth of the current label set.
We then contract ImageNet label nodes along the shortest path to construct superclasses.
Specifically, we investigate depths 2, 4, 5, 6, and 7, which result in class counts of 5, 17, 37, 85, and 232 respectively, in order to provide good coverage across a range of label granularities.
Similarly, on iNaturalist, we use the superclass information that comes with the dataset to collapse the label space from 5,089 fine-grained classes to 13 coarse classes. 

For ImageNet pre-training, we find that using the full 1000 classes provides the most robustness.
However, when the label set size is reduced by four times (i.e., taking 232 superclasses at depth 7), model robustness only decreases slightly. From then on, reducing the label set further to 85 classes (depth 6), and then 37 classes (depth 5), does not deteriorate the linear trend further.
Only when we experiment with 17 classes (depth 4) do we find another noticeable reduction in effective robustness. With 5 superclasses as the only pre-training labels (depth 2), pre-trained models are still significantly more robust than models trained from scratch. 

On iNaturalist, we also observe a similar downward shift in linear trend when we reduce the initial label space to its phylum. Refer to Figure~\ref{fig:max_depth} for more details. Overall these findings suggest that using fine-grained labels during pre-training is better for learning representations that are robust to distributions shifts in downstream tasks. But even if only coarse labels are available, pre-training with enough data still has significant robustness benefits to offer. We note that when it comes to absolute F1 scores, pre-training on more fine-grained labels also tends to lead to higher performance.

\subsection{Effect of Label Semantics}
The number of pre-training classes seems to have an impact on downstream robustness, but does it matter what kind of classes models are pre-trained on? We next investigate whether using classes whose semantics are more aligned with the downstream task would improve robustness. 

\begin{figure}[]
\begin{minipage}{0.45\textwidth}
\includegraphics[trim=0 0 0 0,clip,width=\linewidth]{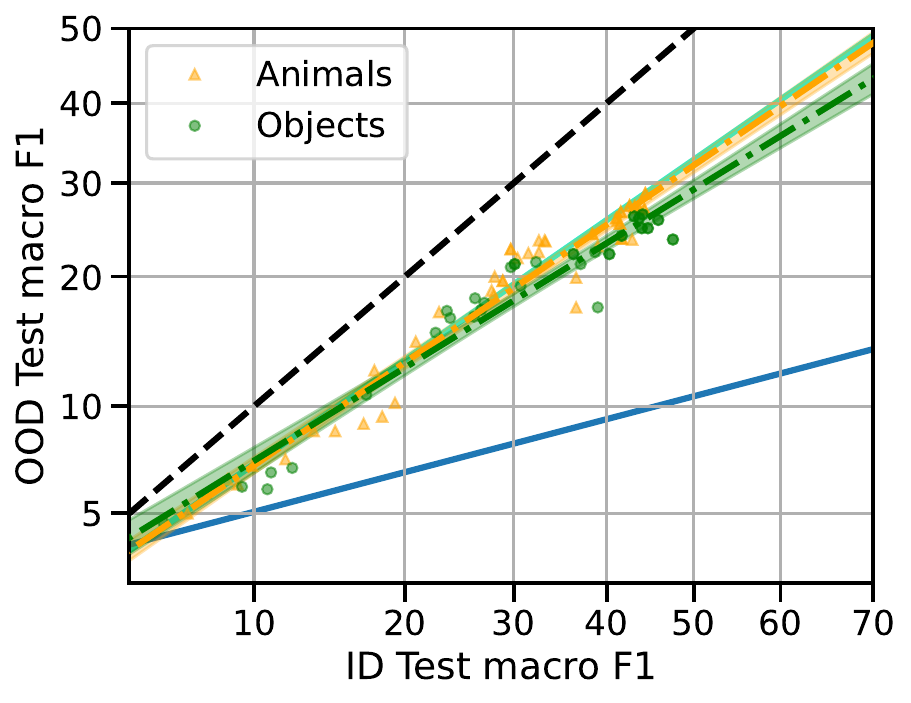}
\end{minipage}\hfill
\begin{minipage}{0.5\textwidth}
\caption{Varying the semantic category of classes included in the pre-training data yields similar robustness linear trends, with pre-training on only ``animal'' classes exhibiting slightly higher F1 scores than pre-training on only ``object'' classes. Even though the downstream task is animal classification, models pre-trained only on ``object'' classes are still much more robust than models that do not undergo any pre-training.
}
\label{fig:animal_object}
\end{minipage}
\vspace{-0.5em}
\end{figure}

To do so, we separately pre-train models on ImageNet classes that are subsets of the ``object'' and ``animal'' WordNet synsets. This yields 2 broad categories that are similar in total sample size, each having around 600K images. In Figure \ref{fig:animal_object}, we find that models trained on ``animal'' classes (yellow line) exhibit slightly higher F1 scores, but roughly the same effective robustness as models trained on "object" classes (green line).
This is surprising given that the fine-tuning distribution, iWildCam, contains only images of animals in the wild, which are semantically more similar to the animal classes of ImageNet. It is also worth noting that models pre-trained on ``object'' classes are also much more robust than models trained from scratch (blue line).

One potential confounder to this experiment setup is that some images from ``object'' classes also contain animals (i.e., co-occurrences that are not accounted for by ImageNet labels). To estimate the extent of this problem, we use TensorFlow’s ImageNet2012 multilabel set \cite{shankar2019evaluating}, containing 20k ImageNet validation data with multi-class labels reviewed by experts. We find that 1.1\% of the data have labels from both ``animal'' and ``object'' classes present in the same image, suggesting that the label co-occurrence issue only impacts a small fraction of training data. Consequently, to explain the result of Figure \ref{fig:animal_object}, we posit that training on a diverse set of classes in general helps the model pick up on useful invariances that in turn lead to similar downstream robustness. We explore this hypothesis further in Section \ref{subsection:inat_pre-training} with synthetic training data. 
\subsection{Effect of Image Diversity}
Besides labels, another source of diversity comes from the training images themselves. We experiment with two different notions of image diversity: \textbf{(i)} Number of categories collected, and \textbf{(ii)} Per-class image diversity.
\subsubsection{More Data Per Class vs. More Classes of Data}\label{sec:label_diversity}
\begin{figure}
\begin{minipage}{0.45\textwidth}
\includegraphics[width=\linewidth]{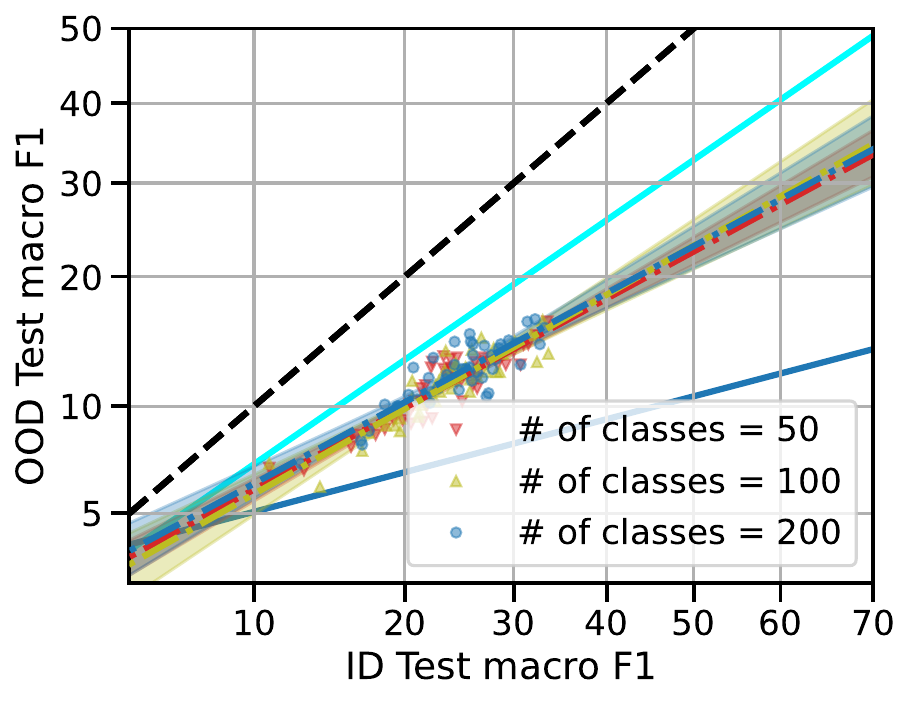}
\end{minipage}\hfill
\begin{minipage}{0.5\textwidth}
\caption{We vary the number of classes randomly selected from the original 1000 ImageNet classes and adjust the number of images per class correspondingly, such that total image quantity is fixed at 60K. We observe that having 4$\times$ more classes, or 4$\times$ more images per class, induces the same level of robustness in fine-tuned models. Experiments with 300K data regime can be found in Appendix Figure \ref{fig:fixed_size_300k}.
}
\label{fig:fixed_size_60k}
\end{minipage}
\vspace{-0.5em}
\end{figure}

A natural question arises when designing a dataset with a fixed data budget (or labeling cost): \textit{should we collect more data from existing categories or more categories of data?} To address this question, we keep the total number of images fixed while varying the number of classes of ImageNet used for pre-training. For example, if we have a budget of 60K images and 100 randomly selected ImageNet classes, we sample 600 images uniformly at random from each of these classes (Figure \ref{fig:fixed_size_60k}).
Here, we find that in the context of the iWildCam distribution shift, there is no difference on downstream robustness between having more data per class or having more classes, as long as the total number of images is constant. This observation also holds at a larger data quantity scale (300K images, see Appendix Figure~\ref{fig:fixed_size_300k}). This result demonstrates the dominant effect of data quantity over other aspects of the pre-training distribution (e.g., label set size).
\subsubsection{Image Diversity Within Each Class}
\label{sec:per-class-div}
Another way of modulating dataset diversity is by changing \emph{per-class} image diversity. For example, given a ``dog'' class, a dataset which only contains images of one dog breed could be seen as less diverse than a dataset which has examples of several breeds. In order to construct a controlled experiment, we use a quantitative heuristic for the diversity of each class: we fix certain superclasses (using the WordNet hierarchy) as the training labels and vary the number of corresponding subclasses where the images are taken from. For iNaturalist we can do the same with the tree of life structure.
More diversity means more subclasses chosen per superclass.

\begin{figure}

\centering
{\includegraphics[width=0.4\textwidth]{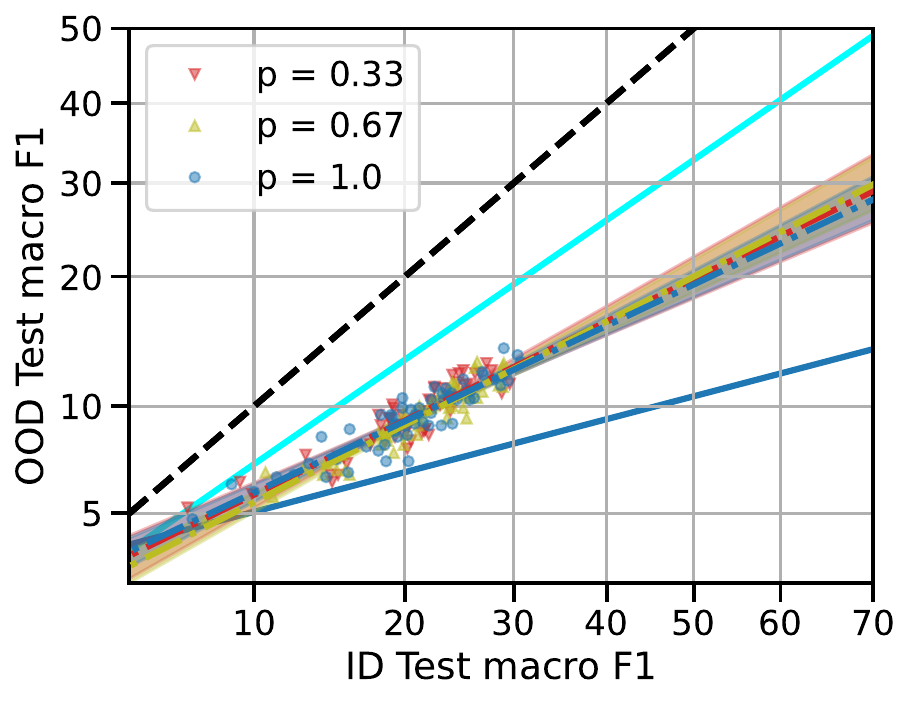}}
{\includegraphics[width=0.4\textwidth]{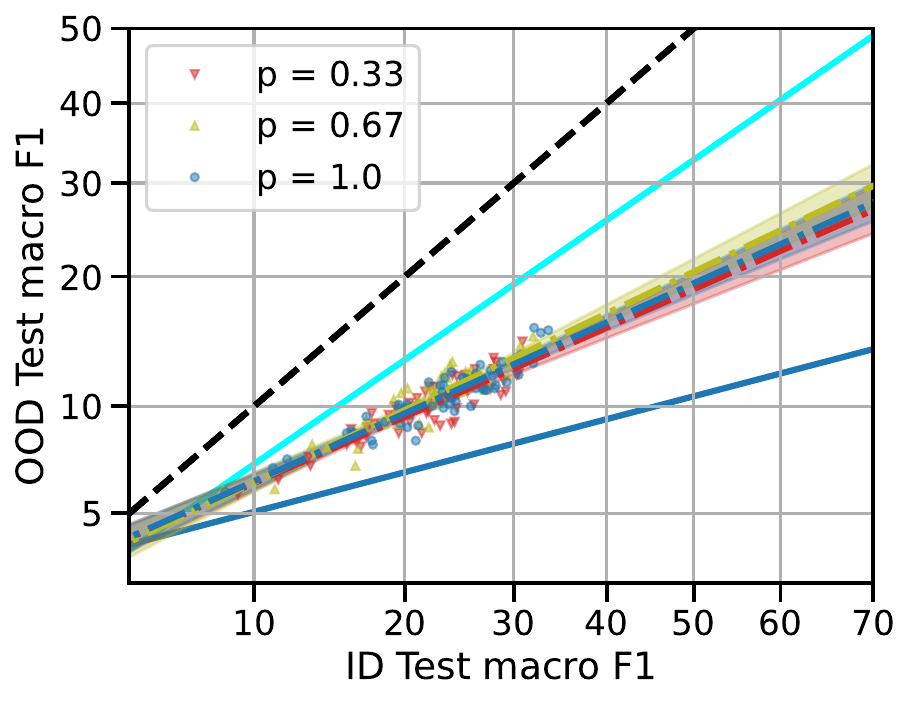}}
\vskip -.75em
\caption{We fix the total amount of pre-training data and the label space, while reducing the number of subclasses that constitute each superclass label in \textbf{(left)} ImageNet and \textbf{(right)} Naturalist. Smaller $p$ (diversity ratio) means proportionally fewer subclasses per superclass. We find that reducing per-class diversity by up to 3$\times$ during pre-training has no effect on downstream robustness.}
\label{fig:image_diversity}
\end{figure}

For the ImageNet distribution that is built on the WordNet hierarchy, we construct a subset following BREEDS~\cite{santurkar2020breeds}.
The two main ways that BREEDS recalibrates the WordNet hierarchy fit our goals for image diversity:
(i) nodes are only included in the hierarchy if they convey some visual information, and
(ii) nodes of similar specificity share the same distance from the root (e.g., ``dog'' and ``cat'' are now both at the same depth even if the ``dog'' subtree is much larger).
With this new hierarchy, we obtain 16 superclasses, each encompassing 12 subclasses (i.e., original ImageNet classes).
The full list can be found in Appendix~\ref{app:image_diversity}. We vary image diversity by changing the number of subclasses per superclass: 4, 8 or 12 subclasses---corresponding to the diversity ratio $p=0.33, p=0.67, p=1.0$ in the left panel of Figure \ref{fig:image_diversity}. To prevent data quantity from being a confounding variable, we subsample images from each chosen subclass accordingly (e.g., if we reduce number of subclasses per superclass by 3$\times$ then we sample 3$\times$ more images from each subclass).

For iNaturalist, we fix the total number of images at 80K and apply the same procedure described above to select a fraction of subclasses (see diversity ratios in the right panel of Figure~\ref{fig:image_diversity}), for each of the following superclasses: ``Plantae'', ``Insecta'', ``Mammalia'', ``Fungi'', ``Aves'', ``Reptilia'', and ``Amphibia.'' We choose this set of superclasses so we could have a uniform distribution of images per class while maintaining the same number of images as our ImageNet experiment. For more details, see Appendix \ref{app:image_diversity}.
As seen in Figure \ref{fig:image_diversity}, for both ImageNet and iNaturalist, the resulting linear trends are highly similar regardless of the diversity ratio $p$, or the number of subclasses per superclass. We conclude that in this case, per-class image diversity does not have a significant impact on downstream robustness. Note that this does not hold in the extreme setting, e.g. repeating the same image to produce a dataset. 
\subsection{Pre-training with Different Data Sources}\label{sec:data_sources}
\label{subsection:inat_pre-training}
\begin{figure}[t]
\centering
{\includegraphics[width=0.24\textwidth]{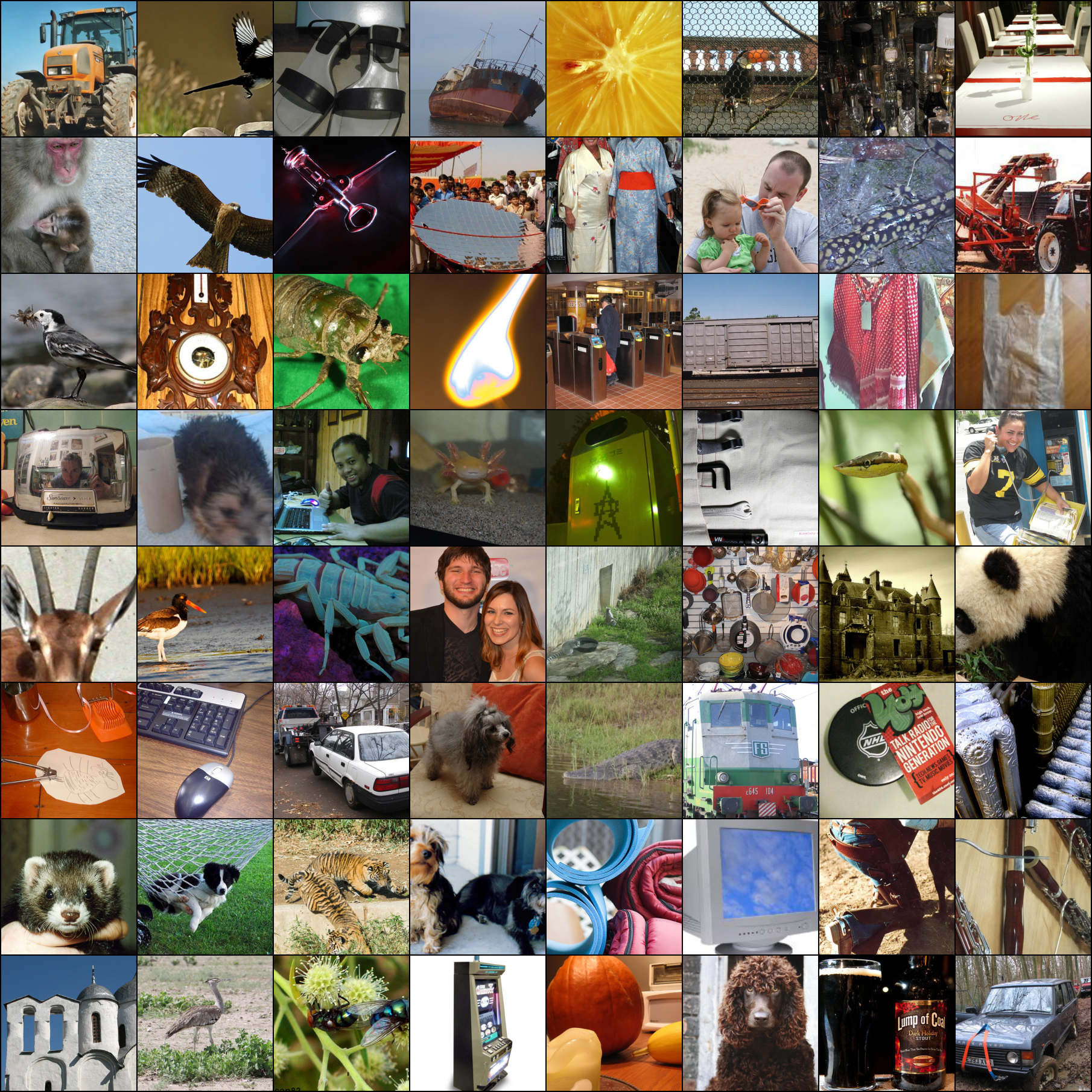}}
{\includegraphics[width=0.24\textwidth]{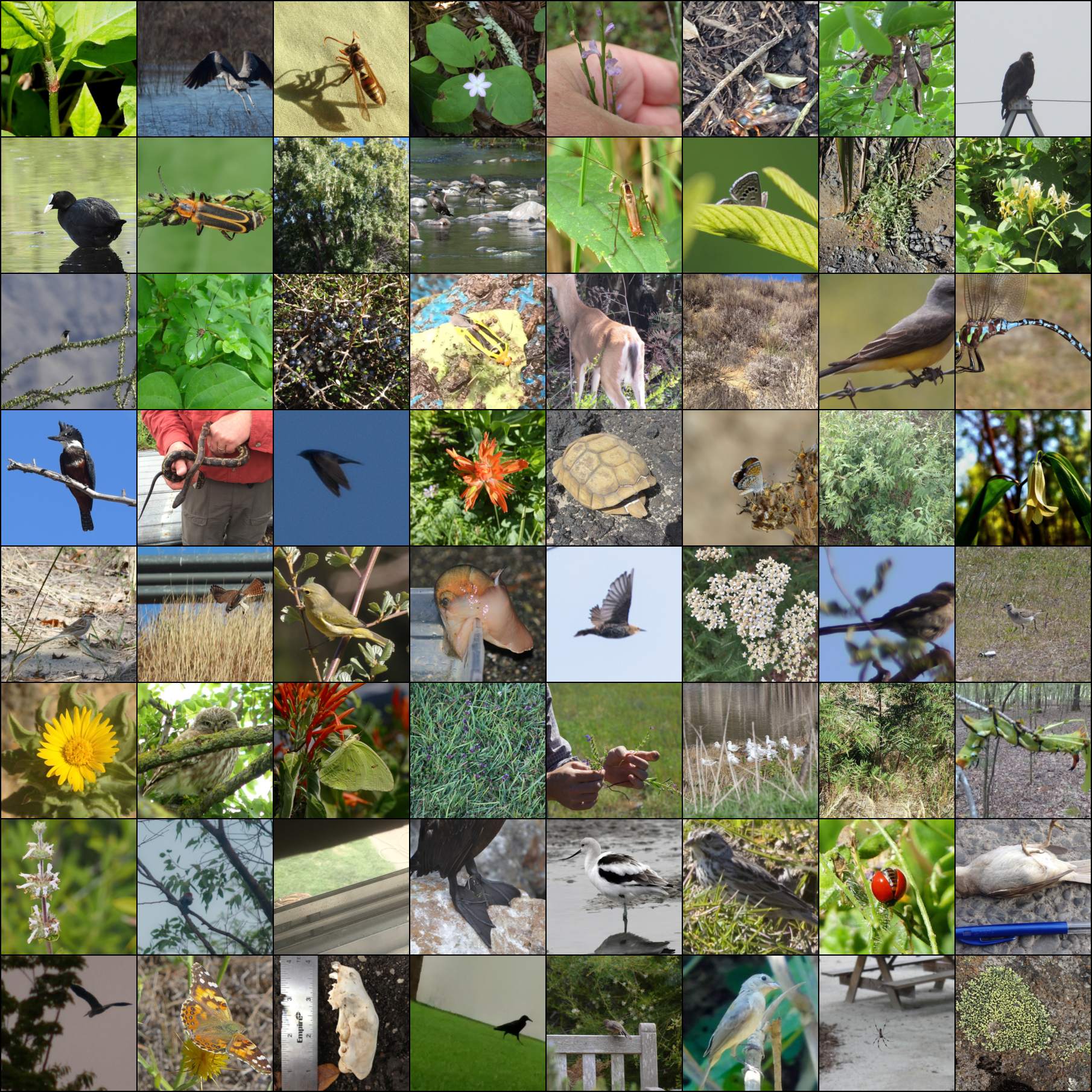}}
{\includegraphics[width=0.24\textwidth]{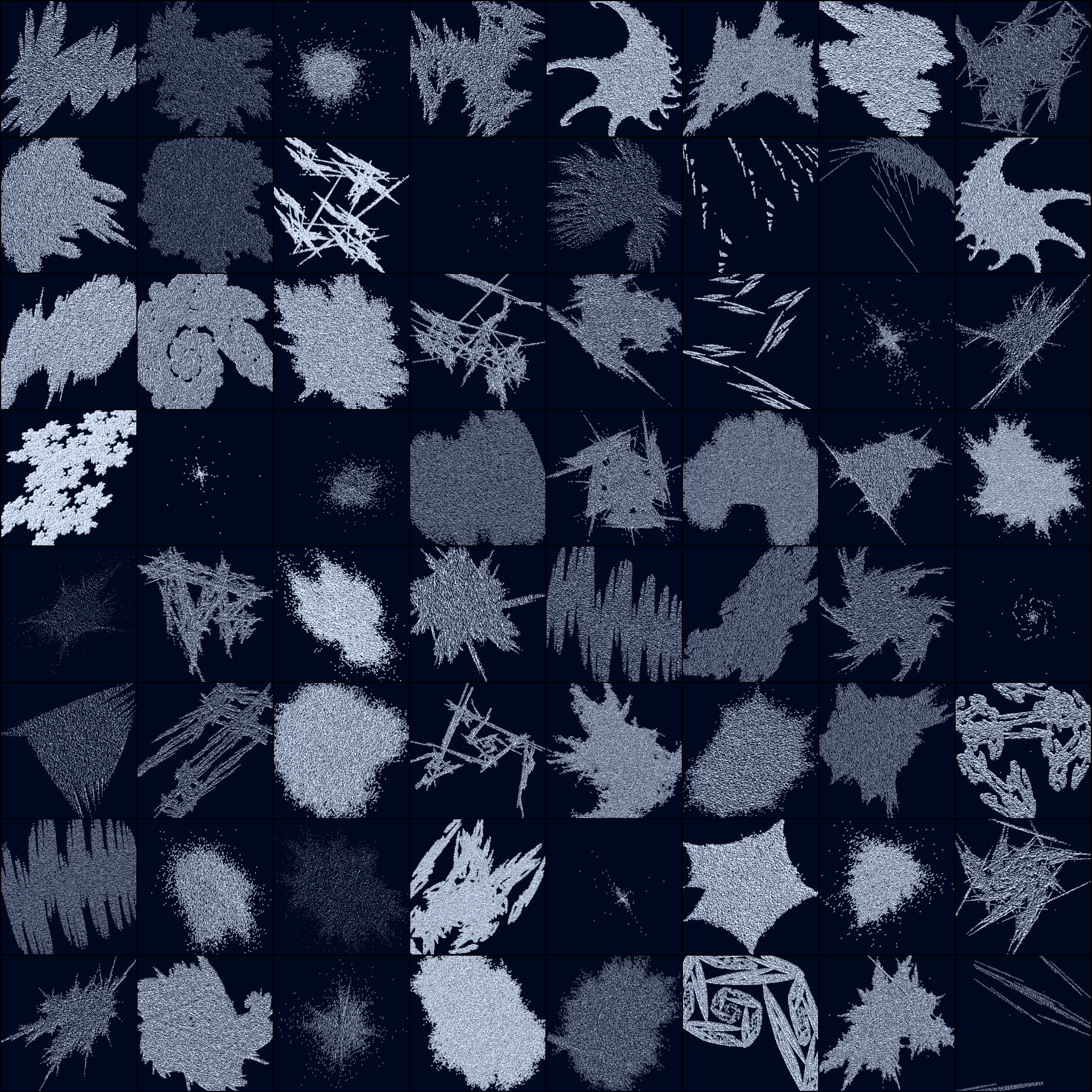}}
{\includegraphics[width=0.24\textwidth]{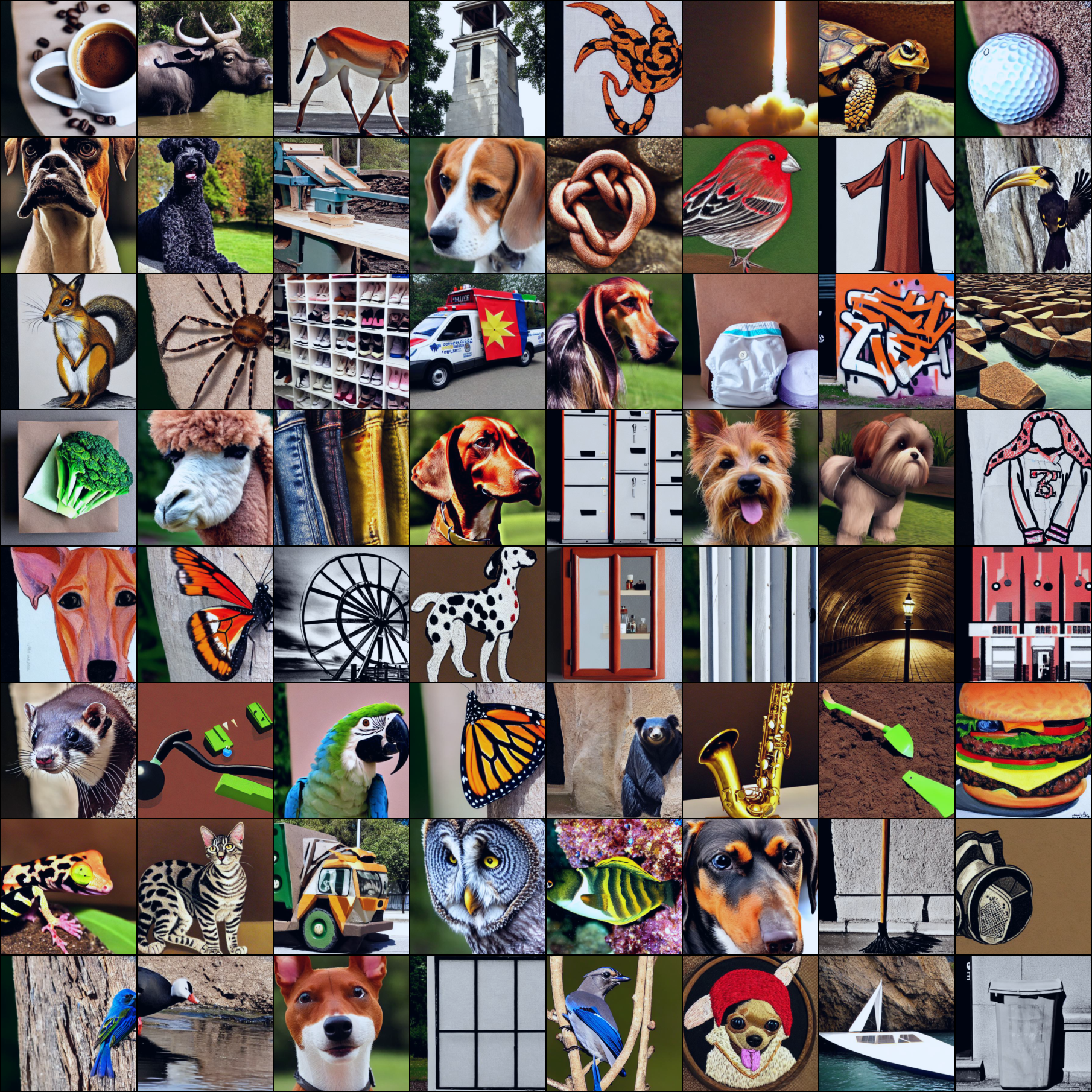}}
\vskip -.25em
\caption{Each grid in order shows random examples from the ImageNet ILSVRC 2012 challenge train set~\cite{russakovsky2015imagenet,deng2009imagenet}, the iNaturalist 2017 challenge train set~\cite{van2018inaturalist}, the FractalDB-1k synthetic train set~\cite{KataokaIJCV2022}, and a 1000-class ImageNet-style synthetic dataset generated using Stable Diffusion \cite{rombach2021highresolution}.}
\label{fig:imagenet_iwildcam_trainsets}
\end{figure}

Moving beyond interventions \textit{within} each data distribution, we now compare fine-tuning robustness behaviors \textit{across} different pre-training data sources.

Compared to ImageNet, iNaturalist exhibits different characteristics on multiple axes (see Table \ref{tab:inat_imagenet}). 
We expect that pre-training on the diverse, domain-specific species -- which have been verified by nature enthusiasts -- in iNaturalist will provide a boost on robustness for the downstream animal-in-the-wild classification task, compared to training on general web-curated classes in ImageNet. However, in Figure \ref{fig:inaturalist_imagenet_fractal}, we find that iNaturalist behaves similarly to ImageNet as a pre-training data source. Even when we subsample iNaturalist to follow the ImageNet class distribution (refer to Section \ref{sec:data_quantity} for its construction), we observe a similar level of effective robustness compared to the equal-sized 150K ImageNet subset (Figure \ref{fig:data_source_150k}). We hypothesize that when a certain level of ``diversity'' is reached with the training images and labels, there is negligible robustness gain to be made even if we increase the alignment between the pre-training and fine-tuning data domains.

\subsubsection{Synthetic Data Sources}
\begin{figure}
\begin{minipage}{0.45\textwidth}
\includegraphics[trim=0 0 0 0,clip,width=\linewidth]{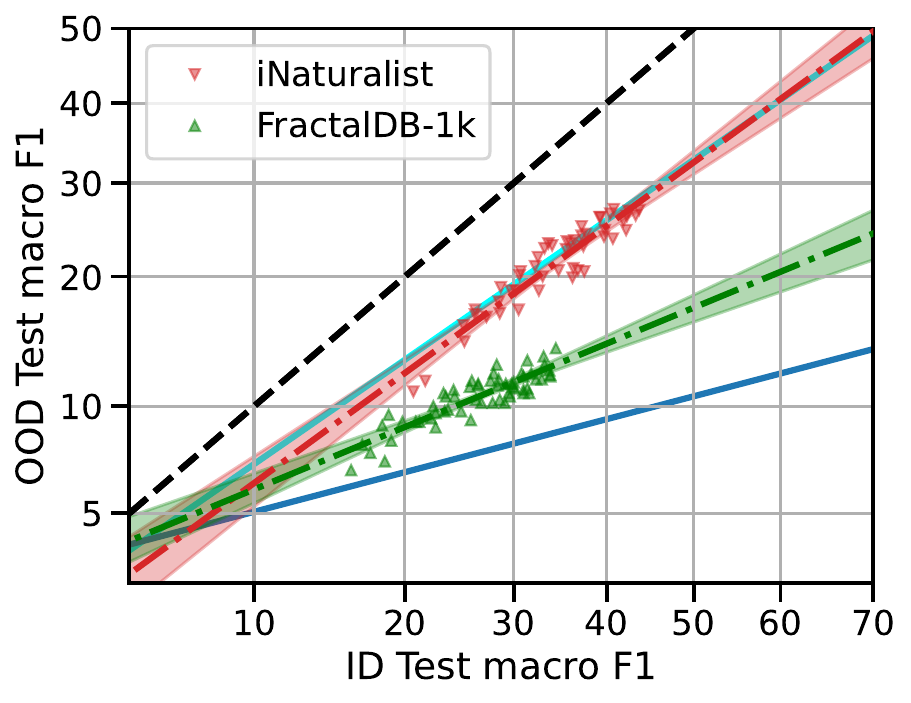}
\end{minipage}\hfill
\begin{minipage}{0.5\textwidth}
\caption{Pre-training on a noisy, long-tailed distribution of natural images like iNaturalist (red line) does not change the robustness on downstream task, compared to pre-training on a clean, class-balanced dataset like ImageNet (cyan line). Pre-training on the same quantity of synthetic fractal data (FractalDB-1k) yields much lower robustness (green line), but still has some benefits compared to training from scratch (dark blue line).
}
\label{fig:inaturalist_imagenet_fractal}
\end{minipage}
\vspace{-0.5em}
\end{figure}

To push our diversity hypothesis to the limit, we also pre-train the same set of architectures on FractalDB-1k dataset~\cite{KataokaIJCV2022}, which has similar class distribution to ImageNet but only contains synthetic fractal images. Pre-training on FractalDB-1k has been shown to surpass the accuracy of pre-training on ImageNet/Places \cite{KataokaIJCV2022}. For the task of iWildCam-WILDS, however, it is noticeably less effective at improving downstream robustness compared to natural image data (Figure~\ref{fig:inaturalist_imagenet_fractal}). However, pre-training with fractal images still offers more robustness than training from scratch.

\begin{figure}
\begin{minipage}{0.45\textwidth}
\includegraphics[trim=0 0 0 0,clip,width=\linewidth]{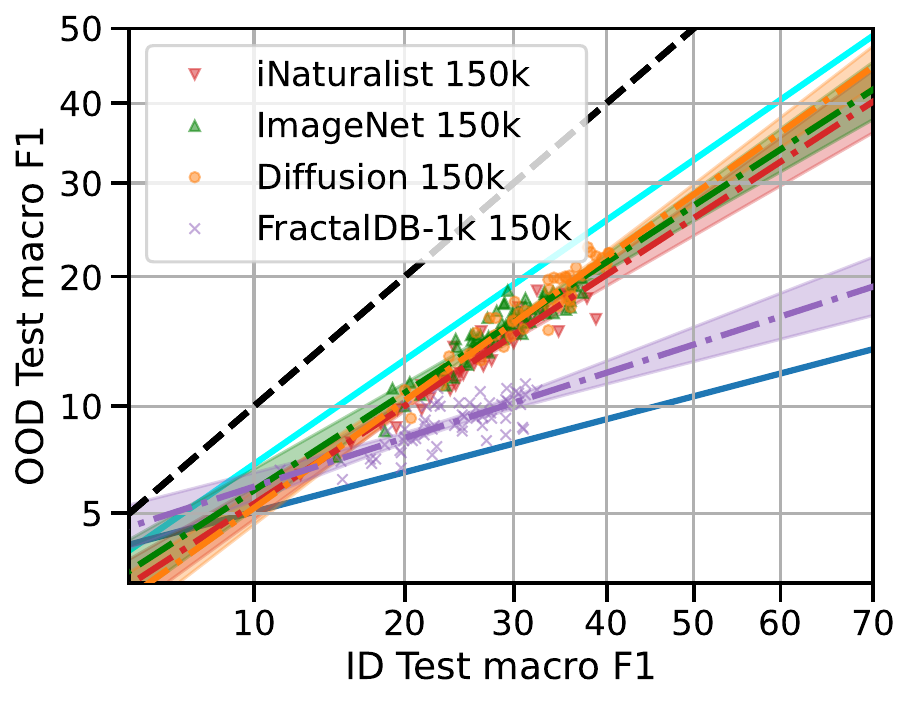}
\end{minipage}\hfill
\begin{minipage}{0.5\textwidth}
\caption{Similar to the results in Figure \ref{fig:inaturalist_imagenet_fractal}, with a budget of 150K images, pre-training on FractalDB-1k is significantly less effective than iNaturalist or ImageNet. However, we find that generating natural-looking data with Stable Diffusion can close this gap. All datasets are subsampled to be class-balanced, each having 1000 classes. We note that at larger pre-training data regime (e.g., 1M samples, see Figure \ref{fig:data_source_1M}), the effectiveness of Stable Diffusion data begins to lag behind natural data from ImageNet and iNaturalist.}
\label{fig:data_source_150k}
\end{minipage}
\vspace{-1.5em}
\end{figure}

Can we generate better synthetic data for pre-training than FractalDB-1k? We experiment with Stable Diffusion~\cite{rombach2021highresolution}, a popular diffusion model which generates high-quality images from natural language prompts, to generate natural-looking images following the ImageNet class distribution. We use 80 diverse prompts per ImageNet class from~\cite{radford2021learning} to generate a 150K ImageNet-like dataset. Examples from this synthetic dataset can be seen in Figure~\ref{fig:imagenet_iwildcam_trainsets}. We find that pre-training on this dataset yields similar robust generalization behaviors as pre-training on the same quantity of ImageNet and iNaturalist images (Figure~\ref{fig:data_source_150k}). However, at a larger scale of 1M images, the robustness benefits of pre-training with synthetic data begins to saturate and slightly lag behind iNaturalist and ImageNet. See Appendix Figure \ref{fig:data_source_1M} for more details.

Overall our findings demonstrate that while nuances in image semantics during pre-training are not important for fine-tuning robustness (e.g., ``animal'' versus ``object'' classes, or iNaturalist versus ImageNet), it is still beneficial to match the general characteristics of the downstream data (e.g., ``natural-looking" images).

\section{Self-supervised Pre-training}\label{sec:self_supervised}
\begin{figure}
\begin{minipage}{0.45\textwidth}
\includegraphics[trim=0 0 0 0,clip,width=\linewidth]{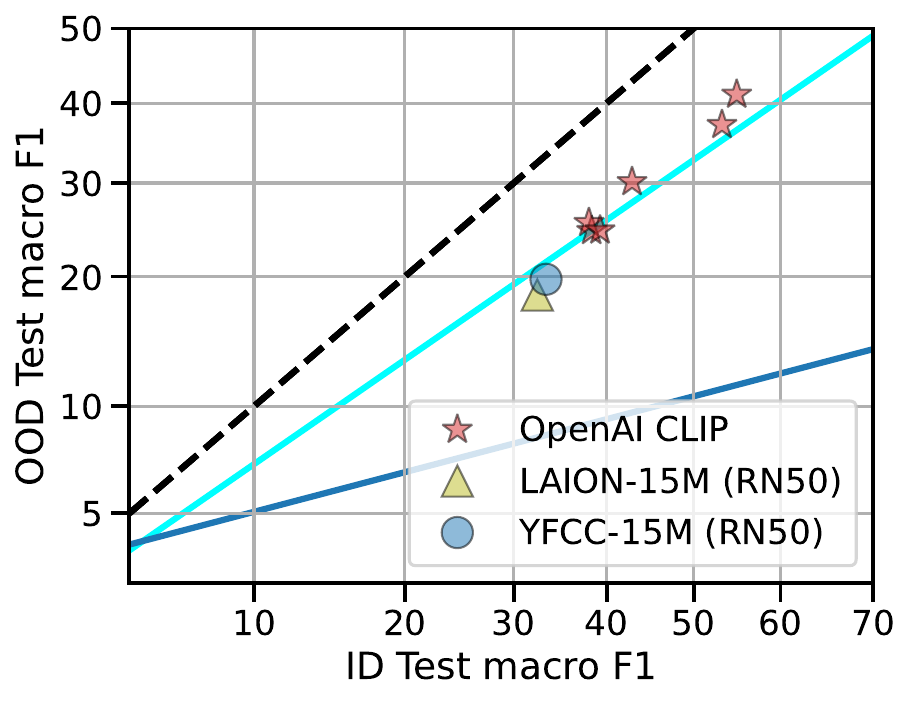}
\end{minipage}\hfill
\begin{minipage}{0.5\textwidth}
\caption{Results from fine-tuning pre-trained CLIP models on iWildCam-WILDS. The models we use include CLIP with ResNet-50 image encoder trained on YFCC-15M \cite{thomee2016yfcc100m} and LAION-15M \cite{schuhmann2021laion} separately, as well as the original CLIP models released by OpenAI \cite{radford2021learning}, including 3 with ViT~\cite{dosovitskiy2020image} backbone trained on a dataset of size 400M. All of these models lie on the same linear trend as that of ImageNet pre-training, demonstrating the consistency of this trend across many pre-training dataset size scales and training algorithms.}
\label{fig:clip}
\end{minipage}
\vspace{-0.5em}
\end{figure}
Previous experiments revolve around the supervised learning settings. However, it is increasingly common to pre-train on {\em self-supervised} tasks using  web-crawled corpora, which has been shown to significantly improve robustness to distribution shifts \cite{radford2021learning}. Our preliminary experiments with pre-trained CLIP models \cite{radford2021learning} on iWildCam show that the resulting ID and OOD performances still lie on the ImageNet pre-training linear trend (i.e., cyan line), despite the self-supervised training mechanism and the much larger training dataset of CLIP. Varying CLIP's data sources only move the F1 scores along the same line (Figure \ref{fig:clip}). 

We also repeat the experiments with varying data quantity (Section \ref{sec:data_quantity}) 
in the context of CLIP's image-text pre-training data. Refer to Appendix \ref{app:clip_additional_experiments} for more details. We leave an extensive evaluation of how ``diversity'' of the pre-training distribution should be defined and measured differently on open-vocabulary web datasets to future work.
\section{Conclusion \& Discussion}
\label{sec:conclusion}
In this work, we find that many factors during pre-training such as label semantics and image diversity, do not significantly alter the effective robustness of models fine-tuned on iWildCam-WILDS. The more influential factors for downstream robust generalization are the \emph{quantity} of the pre-training data and the \emph{granularity} of the pre-training label set. Through experiments with Stable Diffusion, we also demonstrate the potential of synthetic natural-looking images as a way to increase the effectiveness of the pre-training distribution along these two ablation axes.

We can think about pre-training dataset construction in terms of an explore vs.\ exploit trade-off. Exploration, such as finding new data sources, is often time consuming, while exploiting, or collecting as much data as possible from an existing source, can sometimes be significantly easier.
Our experiments suggest that a good approach to building pre-training dataset for robust generalization is to find a few data sources that exhibit robustness characteristics on a downstream task (e.g., Stable Diffusion data), and then collect as many samples from these sources as possible. 

It is important to note that we are studying a different model behavior from \citet{huh2016makes}.
Some interventions can reduce average performance while maintaining effective robustness (e.g., label granularity in Figure~\ref{fig:max_depth}) and vice-versa (e.g., architecture modifications).
Thus, certain choices during pre-training dataset design depend fundamentally on the goals of the dataset.
For example, whether we want to achieve consistent performance across many settings (i.e., robustness) or optimize for very good performance on one specific application changes the applicability of our results.

An important open question is determining what characteristic of the iWildCam-WILDS task leads to the difference in linear trend between pre-training and training from scratch. Some other datasets (e.g., fMoW-WILDS, see \cite{koh2021wilds, fmow2018}) do not exhibit this behavior after fine-tuning. Therefore, it is important to pinpoint distribution shifts where pre-training can provide a significant boost in effective robustness. Finding a unifying property among such datasets would allow for better interpretation of our current results. As a first step in this direction, in Appendix \ref{app:domainnet}, we look into distribution shift settings constructed from the DomainNet benchmark~\cite{peng2019moment}, 
and observe that pre-training and training from scratch only produce different linear trends for certain pairs of domains and not others. Refer to this Appendix for further discussion.

\section*{Acknowledgements}
This work is supported in part by Open Philanthropy, the Allen Institute for AI, and NSF grants DMS-2134012 and CCF-2019844 as a part of NSF Institute for Foundations of Machine Learning (IFML). Ali Farhadi acknowledges funding from the NSF awards IIS 1652052, IIS 17303166, DARPA N66001-19-2-4031, DARPA W911NF-15-1-0543, and gifts from Allen Institute for Artificial Intelligence, Google, and Apple.

\bibliographystyle{abbrvnat}
\bibliography{bibliography}
\clearpage
\appendix
\part*{Appendix}
\vspace{-0.5em}
\section{Training Details} \label{app:train_details}
For the standard deep neural network architectures used in our investigation, their implementation comes from PyTorch's \textit{torchvision.models} package. We use standard hyperparameters found in the WILDS and PyTorch's official GitHub repositories (for training models on iWildCam-WILDS and ImageNet respectively), which match the hyperparameters described in \cite{he2016deep} and \cite{koh2021wilds}. Pre-training on different data sources all uses 90 epochs in total, and then models are fine-tuned on iWildCam for 12 epochs.

\vspace{-0.5em}
\section{Dataset Details}\label{app:data_details}
\begin{figure}[H]
    \centering
    \includegraphics[width=0.55\linewidth]{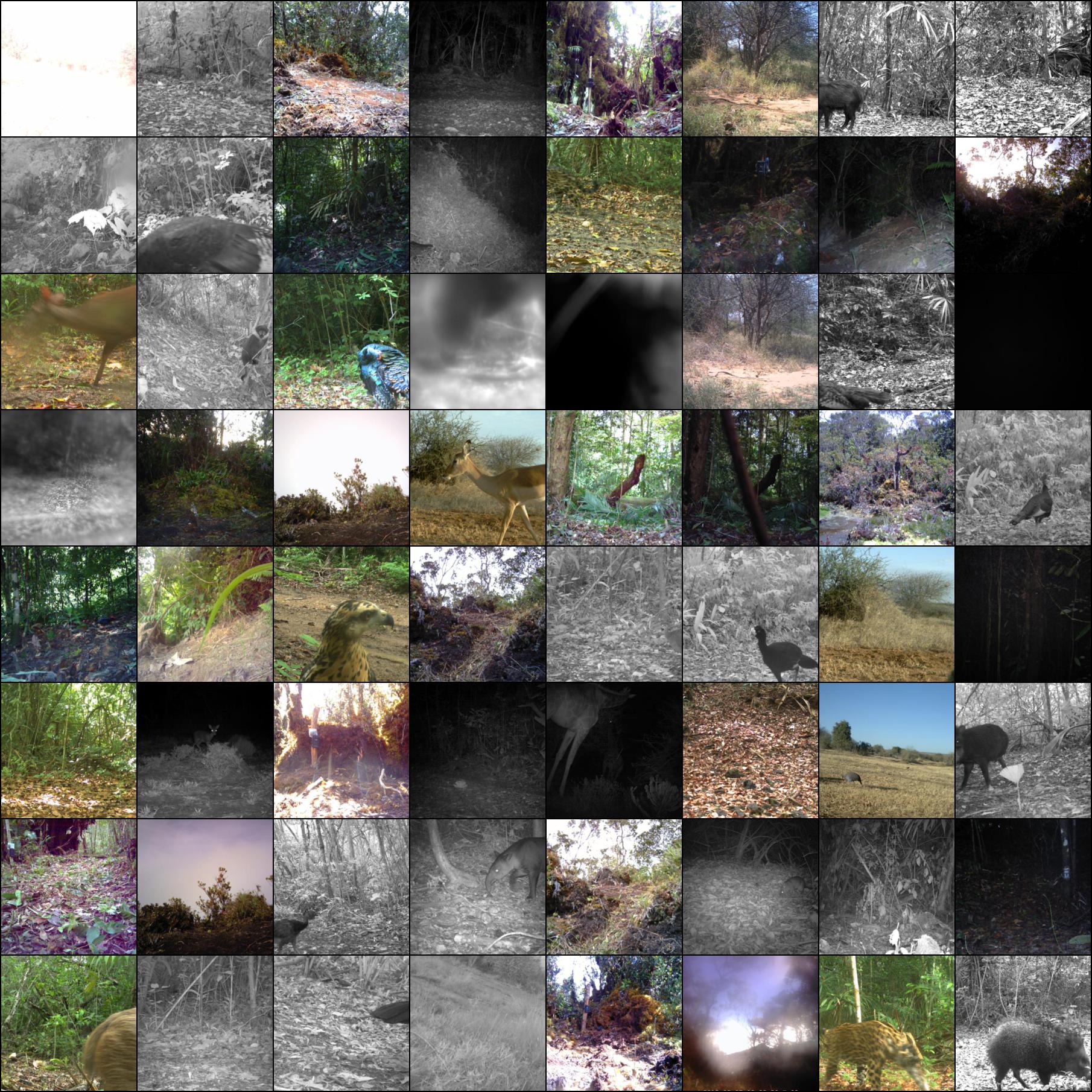}
    \caption{Random examples from the iWildCam-WILDS train set~\cite{koh2021wilds}.}
    \label{fig:iwildcam}
\end{figure}

\paragraph{iWildCam-WILDS}~\cite{biggio2018wild} The iWildCam dataset consists of images of 182 animal species, which are captured through the use of camera traps. We use the iWildCam version 2.0 released in 2021 as a correction to the iWildCam 2020 competition dataset \cite{beery2020iwildcam} to prevent test set leakage. To construct a natural distribution shift, we follow the split proposed by \citet{koh2021wilds}, which results in 2 test sets for evaluation: ID test data consists of images taken by the same camera traps as the training set, but on different days from the training and validation (ID) images, while OOD test data contains images taken by a disjoint set of camera traps from training and validation (ID) images. The train set consists of 129,809 images. The distribution of animal categories over these images is long-tailed (see \cite{koh2021wilds}). The ID test set consists of 8,154 images and the OOD test set consists of 42,791 images, also not class balanced. Examples of train set images can be seen in Figure~\ref{fig:iwildcam}. 

\begin{figure}[H]
    \centering
    \includegraphics[width=0.55\linewidth]{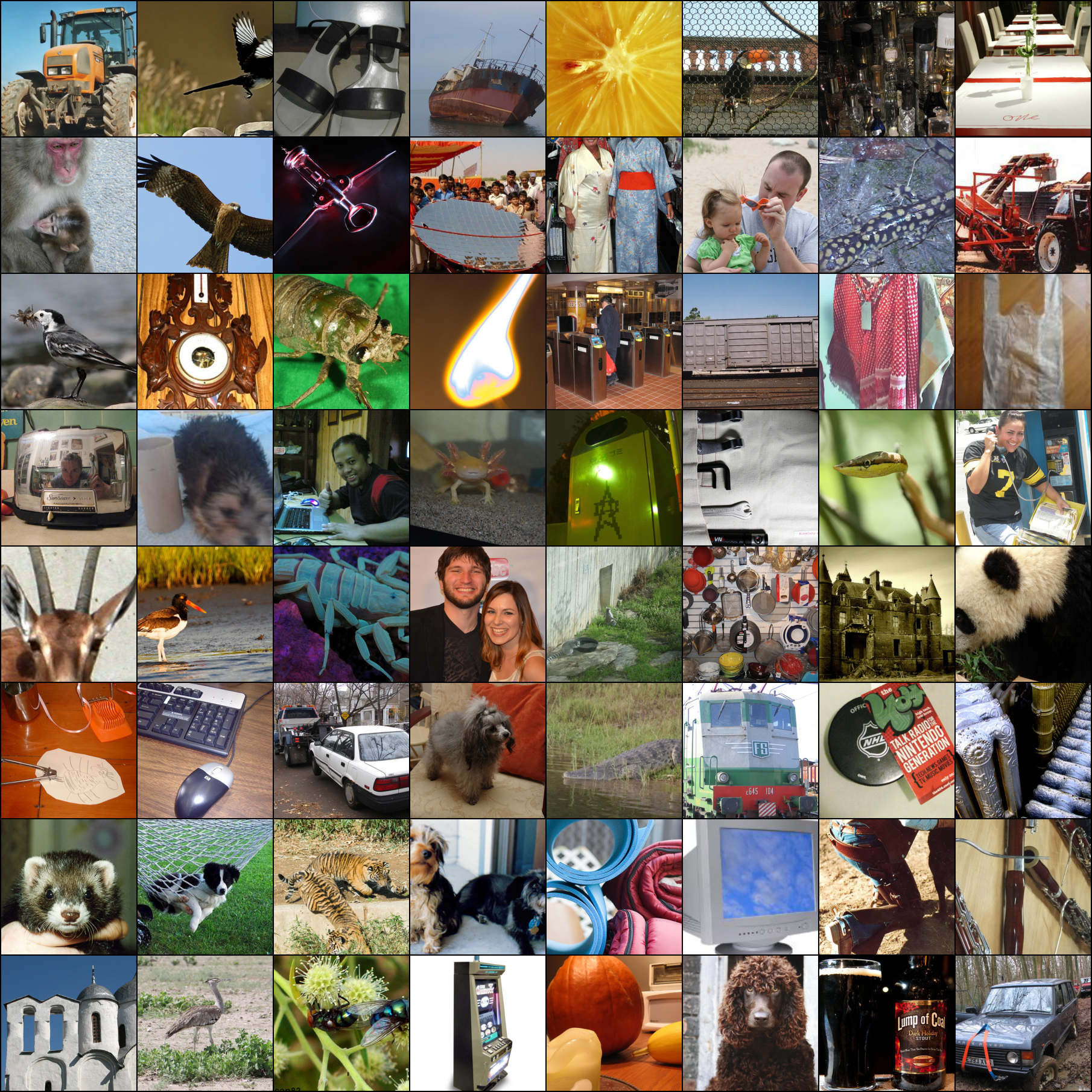}
    \caption{Random examples from the ImageNet ILSVRC 2012 challenge train set~\cite{russakovsky2015imagenet,deng2009imagenet}.}
    \label{fig:imagenet}
\end{figure}

\paragraph{ImageNet}~\cite{russakovsky2015imagenet} We use ImageNet-1k from the ILSVRC 2012 challenge. It contains 1,000 diverse categories of animals and objects, with $\sim$1.2 million training images. The train set is roughly class balanced with $\sim$1.2 thousand images per category. The validation set contains 50,000 images and is exactly class balanced, with 50 images per class. Figure~\ref{fig:imagenet} shows examples of train set images.

\begin{figure}[h]
\centering
{\includegraphics[width=150pt]{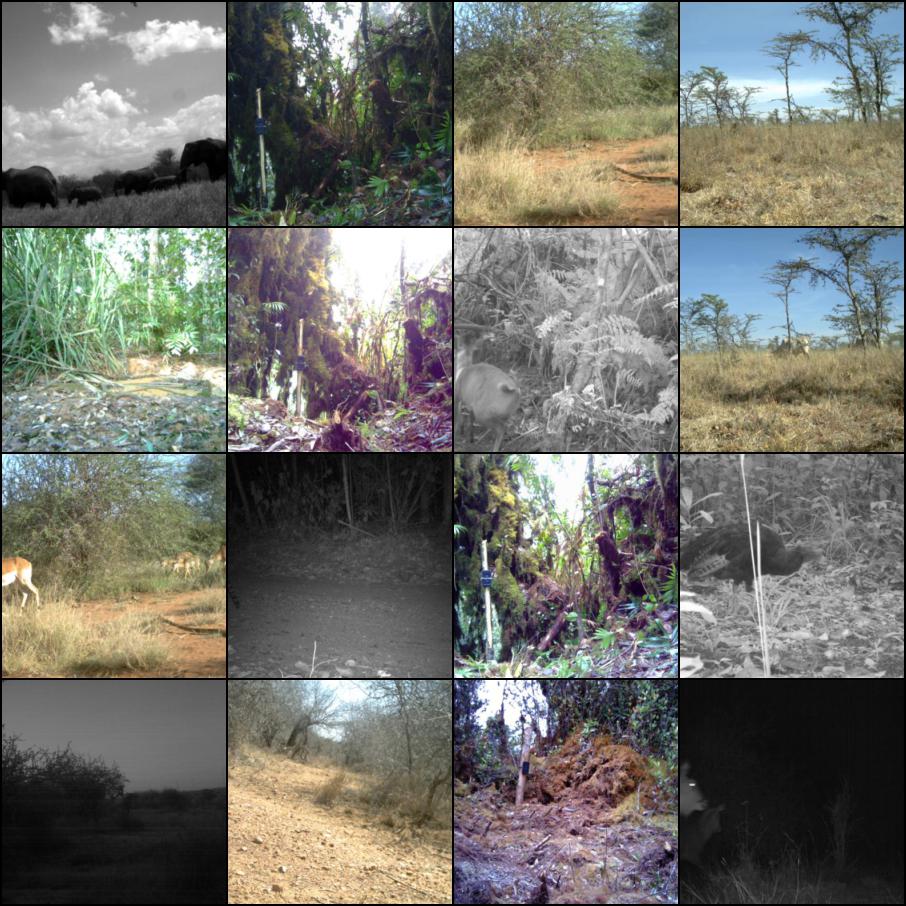}}
{\includegraphics[width=150pt]{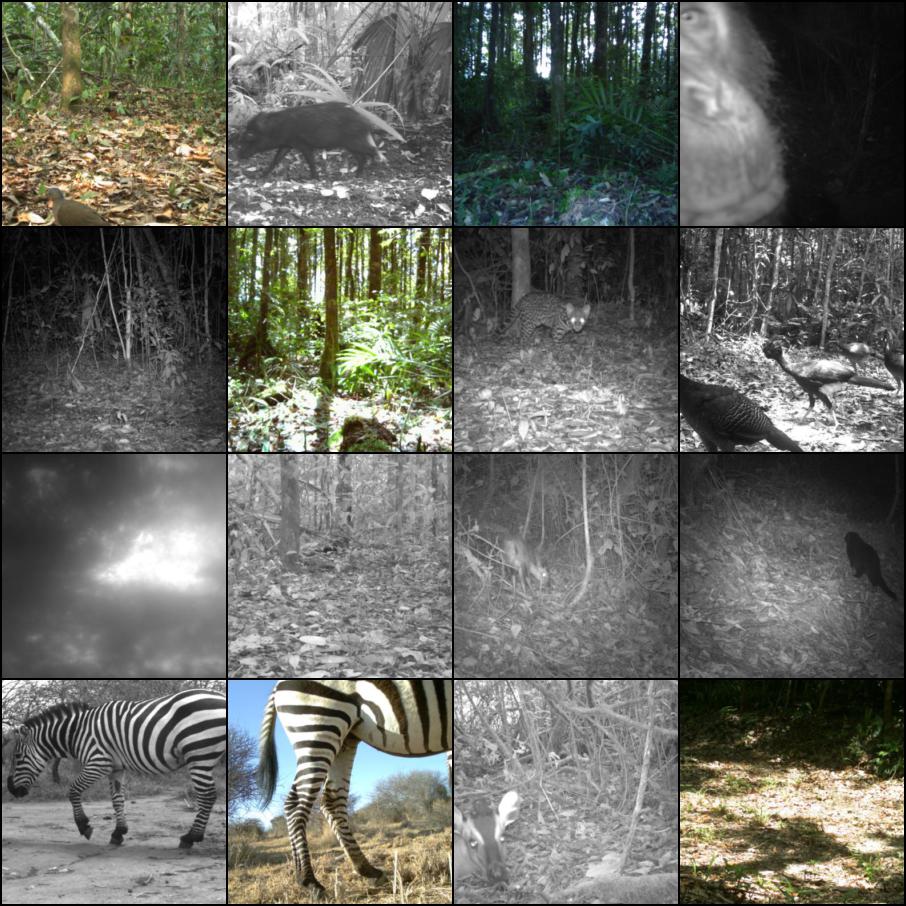}}
\caption{\textbf{(left)} Random images from the in-distribution test set of iWildCam-WILDS~\cite{koh2021wilds}. \textbf{(right)} Random examples from the out-of-distribution test set. This split is based on the geolocations of the camera traps.
}
\label{fig:iwildcam_vals}
\end{figure}

\begin{figure}[h]
    \centering
    \includegraphics[width=0.6\linewidth]{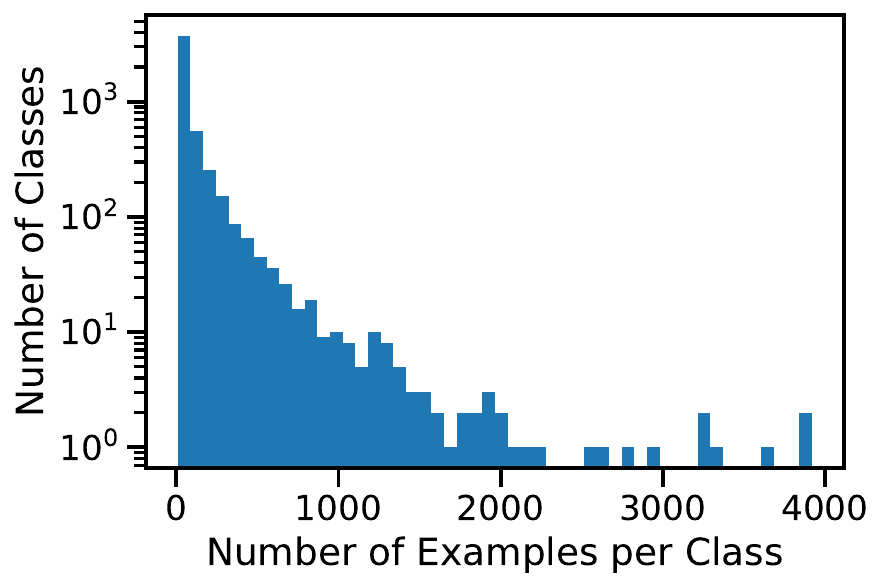}
    \caption{Histogram of class size distribution for the iNaturalist dataset.}
    \label{fig:inat_hist}
\end{figure}

\begin{figure}[H]
    \centering
    \includegraphics[width=0.55\linewidth]{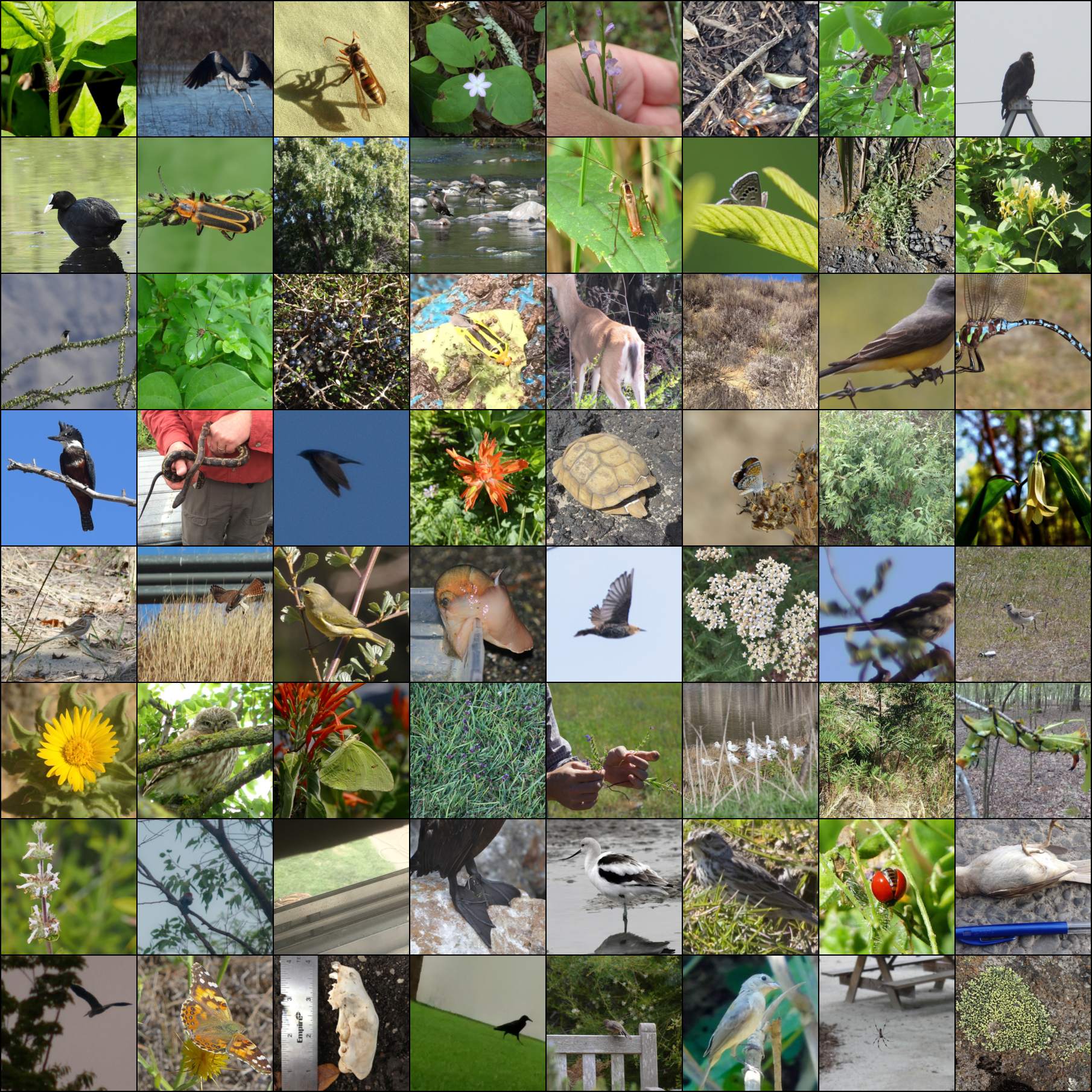}
    \caption{Random examples from the iNaturalist 2017 challenge train set~\cite{van2018inaturalist}.}
    \label{fig:inaturalist}
\end{figure}

\paragraph{iNaturalist}~\cite{van2018inaturalist} We use the version of iNaturalist from the 2017 challenge, with 579,194 training images across 5,089 diverse natural organism categories. The full training set is notably not class balanced, exhibiting a long-tailed distribution (see Figure \ref{fig:inat_hist}). The validation set contains 95,986 images and is also not class balanced, with between 4 and 44 images per category. Refer to Figure~\ref{fig:inaturalist} for examples of iNaturalist data.

\section{Additional Experiment Results}
\subsection{Image Diversity} \label{app:image_diversity}
\begin{figure}[H]
    \centering
    \includegraphics[width=0.5\linewidth]{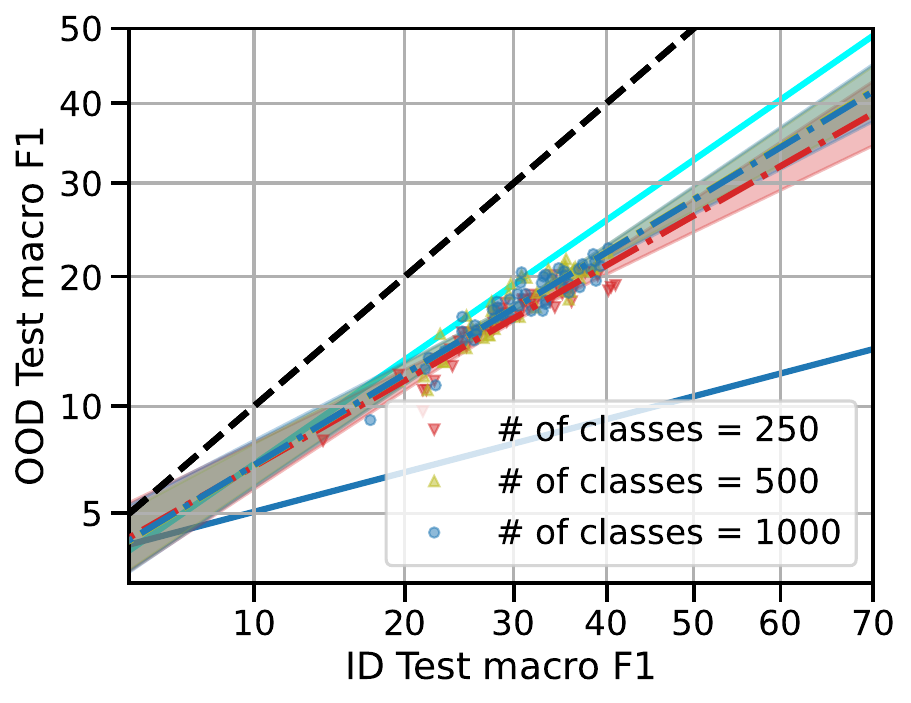}
    \caption{We repeat the experiment in Figure \ref{fig:fixed_size_60k} with a higher data regime. With a total data budget of 300K images, we vary the number of classes randomly selected from the original 1000 ImageNet classes and adjust the number of images sampled from each class correspondingly. We observe that having 4$\times$ more classes, or 4$\times$ more images per class, induces the same level of robustness in fine-tuned models.}
    \label{fig:fixed_size_300k}
\end{figure}

\paragraph{More diverse data or more data per class?} As described earlier in Section \ref{sec:label_diversity}, we vary image diversity by randomly selecting pre-training images from a subset of classes in ImageNet, while keeping the total data budget fixed at 300K. Similar to the findings reported in Section \ref{sec:label_diversity}, in Figure \ref{fig:fixed_size_300k}, we find that varying the number of classes used does not have an effect on the effective robustness of fine-tuned models. 

\paragraph{Per-class diversity.}
For experiments with the BREEDS \cite{santurkar2020breeds} subset of ImageNet, we select 16 superclasses, each having 12 subclasses. The mapping can be found in Table \ref{tab:breeds_image_diversity}. We choose overlapping subsets when modifying our diversity ratio (e.g., the 4 subclasses per superclass chosen when $p=0.33$ are also part of the 8 subclasses per superclass chosen when $p=0.66$). 

For iNaturalist experiments, as mentioned in Section~\ref{sec:per-class-div}, we choose the 7 largest classes so that we could select a uniform number of images per superclass while keeping the total data budget at 80K, matching the setup for ImageNet. When modulating our diversity ratio, we randomly select a subset of subclasses from each superclass and uniformly sample images \textit{with replacement} from these subclasses. As in the ImageNet experiment, subsets of subclasses chosen per superclass are overlapping at increasing diversity ratios.

\begin{table}[]
    \centering
\begin{tabular}{lr}
\toprule
       Superclass Category &  Counts \\
\midrule
      Protozoa &     308 \\
     Chromista &     398 \\
Actinopterygii &    1982 \\
     Arachnida &    4873 \\
      Animalia &    5228 \\
         Fungi &    5826 \\
      Mollusca &    7536 \\
      Amphibia &   15318 \\
      Mammalia &   29333 \\
      Reptilia &   35201 \\
       Insecta &  100479 \\
       Plantae &  158407 \\
          Aves &  214295 \\
\bottomrule
\end{tabular}
    \caption{Superclass data counts for the iNaturalist~\cite{van2018inaturalist} train set. We use the 7 largest classes for our iNaturalist experiment in Section~\ref{sec:per-class-div} so that we could select a uniform number of images per superclass while still having 80K images in total, to match the corresponding ImageNet experiment. Note that some superclass categories are in fact semantic superclasses of other categories (e.g. Aves, Reptilia, and Mammalia are all subclasses of Animalia) but are labeled disjointly. Given that we only use the last 7 categories, our superclass selection avoids this problem.}
    \label{tab:my_label}
\vskip -0.5em
\end{table}

\begin{table*}[h]
\centering
\begin{tabular}{p{0.25\linewidth} | p{0.72\linewidth}}
  \hline
   Superclass & Subclasses \\
  \hline
  garment & trench coat, abaya, gown, poncho, military uniform, jersey, cloak, bikini, miniskirt, swimming trunks, lab coat, brassiere\\
  \hline
  bird & African grey, bee eater, coucal, American coot, indigo bunting, king penguin, spoonbill, limpkin, quail, kite, prairie chicken, red-breasted merganser \\
  \hline
  reptile, reptilian & Gila monster, agama, triceratops, African chameleon, thunder snake, Indian cobra, green snake, mud turtle, water snake, loggerhead, sidewinder, leatherback turtle \\
  \hline
  arthropod & rock crab, black and gold garden spider, tiger beetle, black widow, barn spider, leafhopper, ground beetle, fiddler crab, bee, walking stick, cabbage butterfly, admiral \\
  \hline
  mammal, mammalian & Siamese cat, ibex, tiger, hippopotamus, Norwegian elkhound, dugong, colobus, Samoyed, Persian cat, Irish wolfhound, English setter, llama \\
  \hline
  fish & sturgeon, stingray, coho, great white shark, lionfish, gar, goldfish, hammerhead, rock beauty, anemone fish, barracouta, electric ray \\
  \hline
  accessory, accoutrement, accouterment & shower cap, stole, wig, Windsor tie, bib, necklace, bow tie, gasmask, umbrella, cowboy boot, Christmas stocking, bathing cap \\
  \hline
  appliance & espresso maker, sewing machine, stove, waffle iron, rotisserie, lawn mower, electric fan, dishwasher, iron, microwave, vacuum, space heater \\
  \hline
  craft & gondola, schooner, liner, airship, speedboat, airliner, canoe, trimaran, balloon, lifeboat, yawl, submarine \\
  \hline
  equipment & photocopier, mouse, dumbbell, iPod, home theater, projector, cassette player, hand-held computer, tennis ball, tape player, snorkel, monitor \\
  \hline
  furniture, piece of furniture, article of furniture & bookcase, throne, barber chair, four-poster, desk, medicine chest, crib, chest, sliding door, toilet seat, rocking chair, wardrobe \\
  \hline
  instrument & chime, wine bottle, ladle, reel, wall clock, hammer, wok, abacus, assault rifle, projectile, safety pin, corkscrew \\
  \hline
  man-made structure, construction & lumbermill, scoreboard, monastery, church, tobacco shop, drilling platform, dam, fountain, bell cote, yurt, bookshop, prison \\
  \hline
  wheeled vehicle & convertible, oxcart, electric locomotive, tricycle, fire engine, bicycle-built-for-two, moving van, golfcart, steam locomotive, jinrikisha, tractor, ambulance \\
  \hline
  cooked food, prepared food & consomme, burrito, meat loaf, bagel, ice cream, French loaf, ice lolly, hot pot, cheeseburger, potpie, trifle, mashed potato \\
  \hline
  produce, green goods, green groceries, garden truck & broccoli, cauliflower, mushroom, artichoke, acorn, head cabbage, spaghetti squash, jackfruit, cucumber, orange, hip, banana \\
  \hline
\end{tabular}
\caption{Subclass-superclass mapping for the subset of ImageNet that we use to control per-class diversity, constructed based on the BREEDS hierarchy \cite{santurkar2020breeds}.}
\label{tab:breeds_image_diversity}
\end{table*}
\subsection{CLIP Fine-tuning Details}
We investigate how the effective robustness obtained from ImageNet pre-training would change at a much larger pre-training data regime. To do so, we fine-tune CLIP~\cite{radford2021learning} models that have been trained on different data distributions on iWildCam-WILDS. We use the same finetuning hyperparameters from~\cite{koh2021wilds} with the AdamW optimizer~\cite{loshchilov2017decoupled}. The training datasets for CLIP are often large corpora of image-text pairs scraped from the internet. Models from the original CLIP paper \cite{radford2021learning} are trained with 400M data points, and we analyze specifically those that use ViT \cite{dosovitskiy2020image} as the image encoder architecture. We also include the results from CLIP ResNet50 models trained on YFCC-15M~\cite{thomee2016yfcc100m} and LAION-15M~\cite{schuhmann2021laion} separately, to show the effects of pre-training with different data scales and data sources. The results of these experiments can be found in Figure~\ref{fig:clip}. We find that the linear trend of ImageNet pre-trained models is still predictive of the effective robustness obtained from these CLIP models. Note that along with utilizing significantly more data, CLIP also undergoes a different pre-training mechanism (i.e., contrastive versus supervised learning). This demonstrates the widespread applicability of existing effective robustness trends in analyzing generalization properties of pre-trained models.
\begin{figure}
\centering
\includegraphics[width=0.5\linewidth]{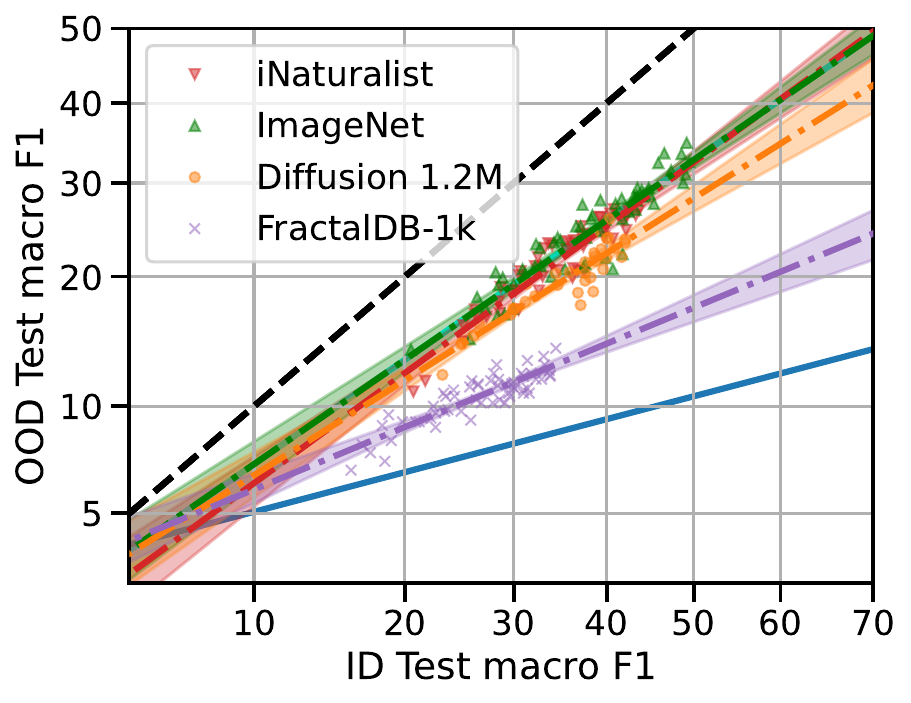}
\vskip -0.5em
\caption{We experiment with pre-training on an ImageNet-scale synthetic dataset (i.e., 1.2M images) generated using Stable Diffusion, and find that it yields less robustness after fine-tuning compared to pre-training on natural data of similar quantities (i.e., ImageNet, iNaturalist). However, the natural-looking synthetic data in Diffusion 1.2M is still much more effective as a pre-training source compared to the synthetic fractal data in FractalDB-1k.}
\label{fig:data_source_1M}
\vskip -0.5em
\end{figure}
\subsection{Pre-training on ImageNet-scale Diffusion Data}
As described in Section \ref{sec:data_sources}, we explore the effect of pre-training on images from different data distributions, both synthetic and natural. At 150K data regime (Figure \ref{fig:data_source_150k}), our earlier findings demonstrate that the transfer robustness obtained from pre-training on natural-looking synthetic data could rival that of pre-training on ImageNet or iNaturalist. Now, we increase the number of synthetic images generated using Stable Diffusion to 1.2M, matching the full ImageNet class distribution. At this data regime, we find that natural-looking synthetic data (i.e., Diffusion 1.2M) is slightly less effective than natural data (i.e., iNaturalist and ImageNet) at improving the effective robustness of fine-tuned models (Figure \ref{fig:data_source_1M}). However, Diffusion 1.2M still offers much more robustness gain compared to pre-training on synthetic fractal data (i.e., FractalDB-1k).

We hypothesize that the additional image diversity obtained from using more synthetic data saturates after certain data quantity, as the objects generated by Stable Diffusion appear to be relatively more centered and less cluttered, compared to ImageNet data (Figure \ref{fig:imagenet_iwildcam_trainsets}). Further study into alternative prompting templates/ techniques is needed to increase the variety of the images obtained from Stable Diffusion.

\section{Experiments with Contrastive Pre-training}
\label{app:clip_additional_experiments}
We attempt to apply our interventions to the constrastive learning regime, done on image-text pairs scraped from the web. In this set of experiments, we construct different pre-training distributions from LAION-5B \cite{schuhmann2021laion}, and evaluate the resulting models \textit{zero-shot} on ImageNet and two ImageNet-derived distribution shifts: ImageNet-V2 \cite{recht2019imagenet} and ImageNet-Sketch \cite{wang2019learning}. We experiment with pre-training CLIP ViT-B/32 on 10M, 50M and 100M samples separately, for 128M steps in total with batch size 4096 per GPU, learning rate 0.0005 and 500 warmup steps. Varying pre-training set size by 10$\times$ from 10M to 100M doesn't significantly alter the robustness linear trends on either distribution shift, despite offering substantial boost in robustness compared to training models from scratch on ImageNet (yellow line, obtained from \cite{miller2021accuracy}).  We hypothesize that much larger scale differences are needed to test the effect of data quantity in this case. This additional investigation is not the focus of this work, however we believe adapting our interventions to massive pre-training webdatasets to be an interesting direction for future exploration.

\begin{figure}[h]
\centering
\includegraphics[width=1\linewidth]{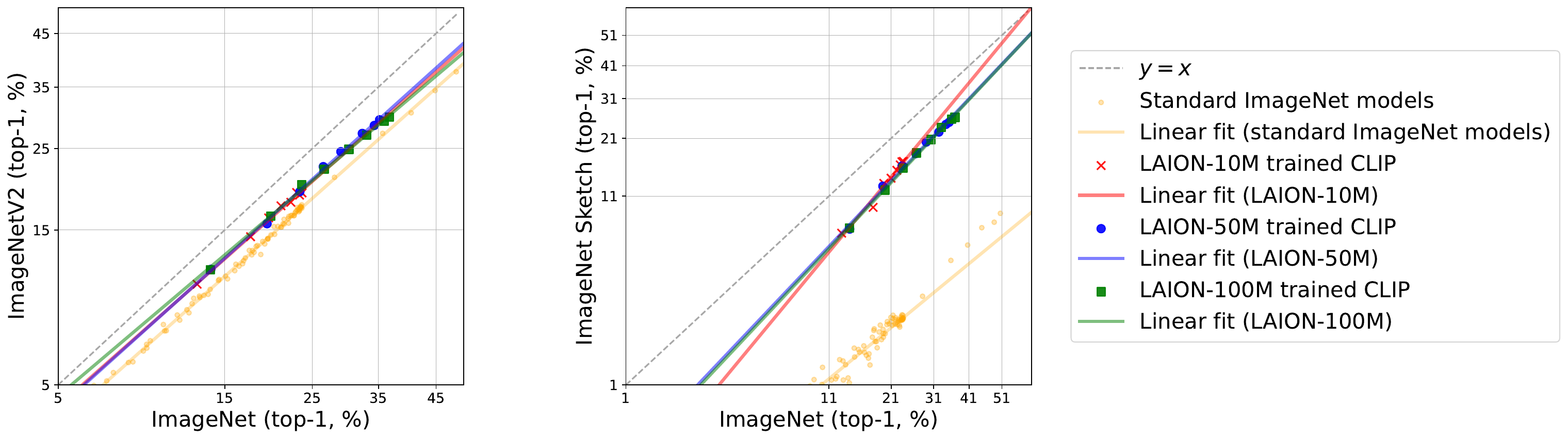}
\vskip -0.5em
\caption{We experiment with pre-training with different data quantities sampled from LAION-5B \cite{schuhmann2021laion}, and find that 10M, 50M and 100M dataset sizes do not change downstream robustness when evaluated zero-shot on ImageNet-derived distribution shifts. However, pre-training still yields much more robustness compared to training from scratch (linear trend obtained from \cite{miller2021accuracy}), especially on the ImageNet to ImageNet-Sketch shift.}
\label{fig:data_quantity_clip}
\vskip -0.5em
\end{figure}

\newpage
\section{DomainNet Experiments}\label{app:domainnet}
\begin{figure*}
    \centering
    \includegraphics[width=\textwidth]{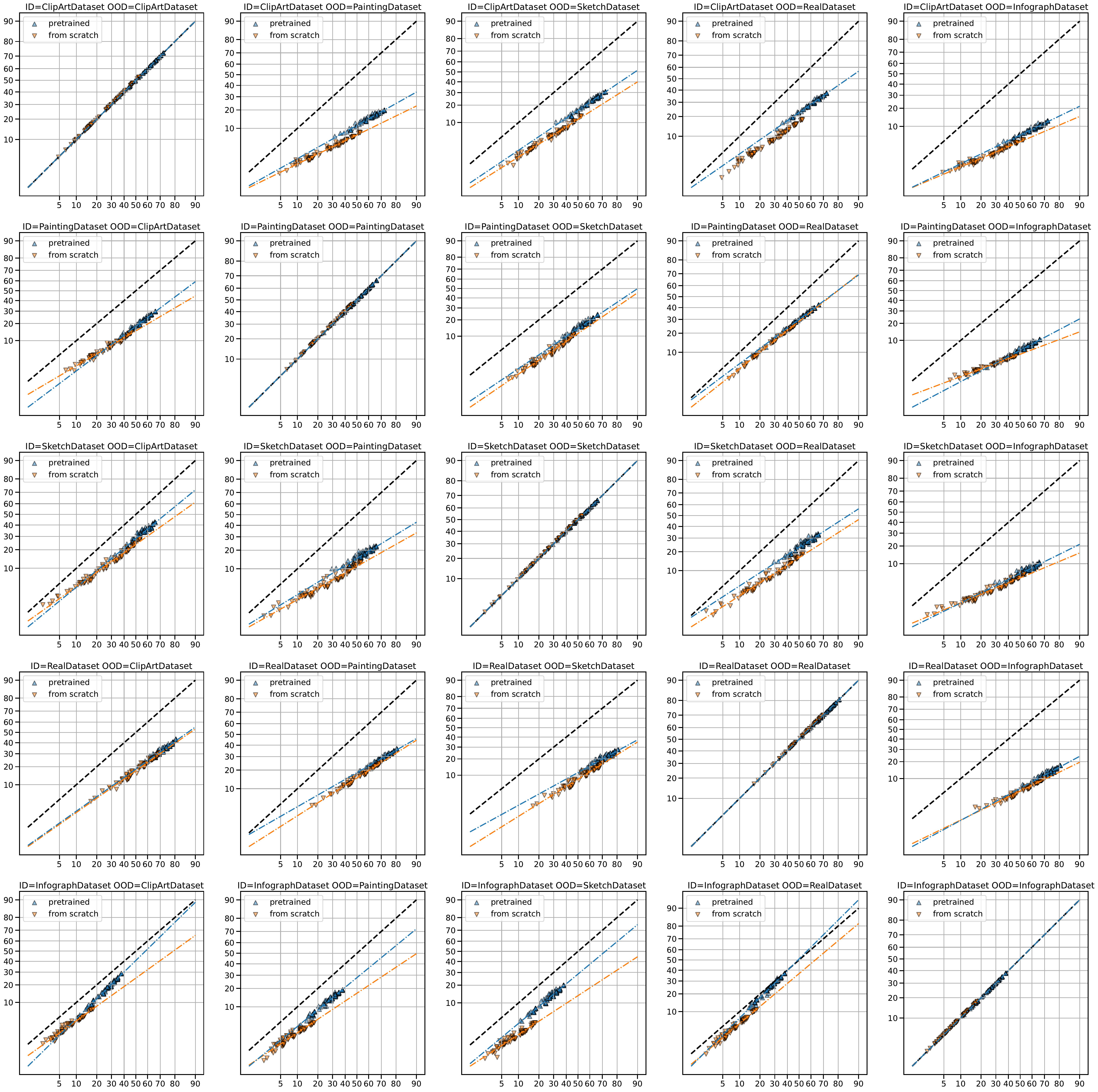}
    \caption{We compare effective robustness of models pre-trained on ImageNet and those trained from scratch on each domain (see each row), using the data from the same domain as ID test set, and data from every other domain as the OOD test set (see each column). In most cases, ImageNet pre-training and training from scratch yield similar downstream robustness, except for the Infograph \& Sketch pair where we observe a sufficiently large gap in effective robustness.}
    \label{fig:domainnet}
\end{figure*}

\begin{figure*}[h]
    \centering
    \includegraphics[width=0.45\textwidth]{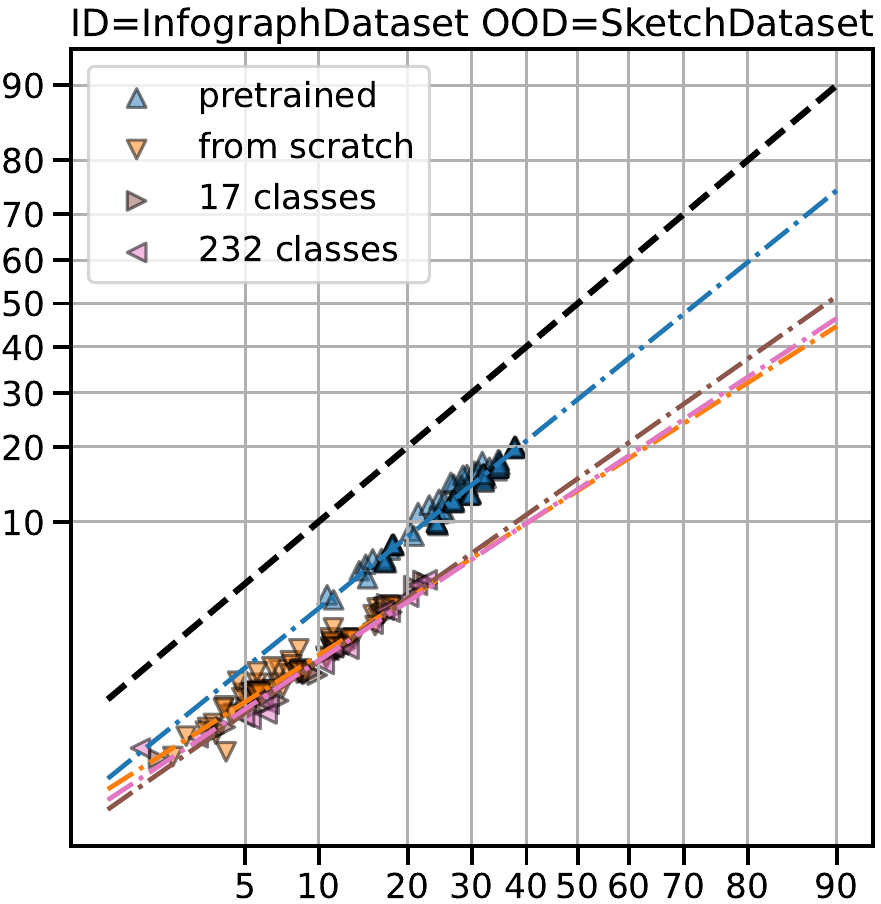}
    \includegraphics[width=0.45\textwidth]{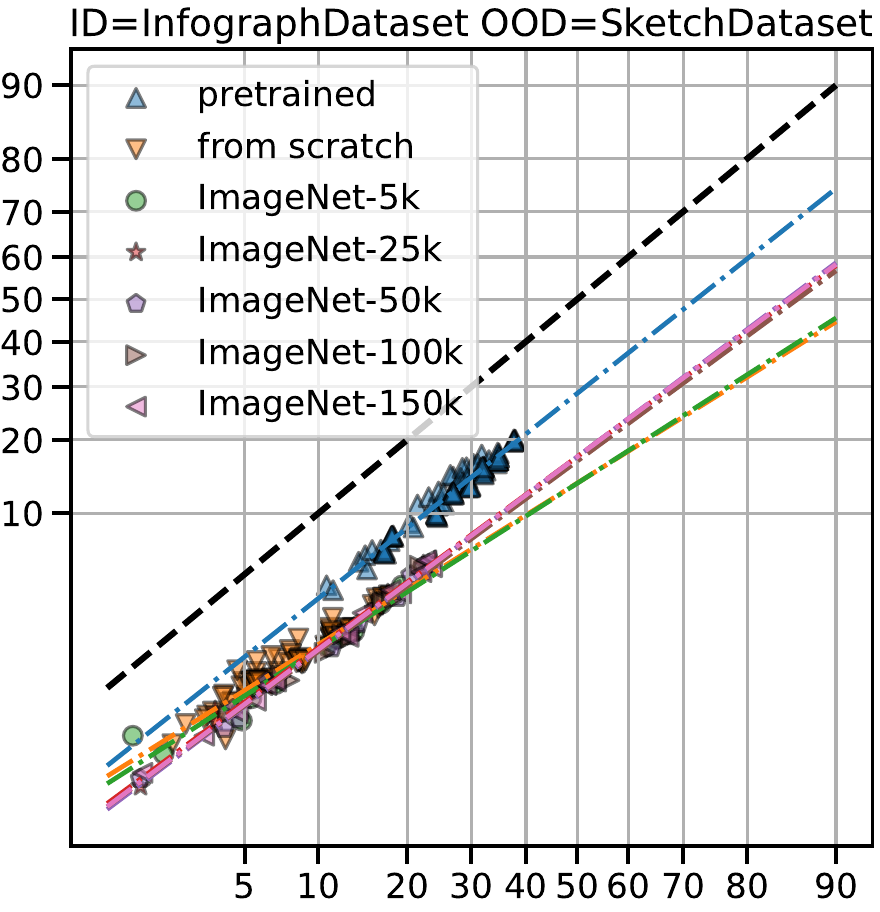}
    \caption{\textbf{(left)} Effect of label granularity. \textbf{(right)} Effect of data quantity. We perform the same interventions described earlier in Sections \ref{sec:data_quantity} and \ref{sec:label_granularity}, on these two aspects of the ImageNet pre-training distribution that have been shown to be influential to downstream robustness. The ID and OOD data comes from Infograph and Sketch domains respectively. We again find that using more coarse labels and smaller data quantity during pre-training lowers the effective robustness of fine-tuned models significantly.}
    \label{fig:domainnet_intervention}
\end{figure*}
For each domain provided in DomainNet \cite{peng2019moment}, we compare models trained from scratch on data from that domain, to models that have been pre-trained on ImageNet and then fine-tuned on the same domain. We then evaluate the model performance on all other domains that have not been used for training, which are considered OOD test sets. Figure \ref{fig:domainnet} shows the resulting effective robustness for all pairs of domains, with $x$ axis and $y$ axis representing ID and OOD accuracies respectively. We find that for most pairs, ImageNet pre-training and training from scratch do not exhibit distinctive linear trends. An exception to this can be found in the case of fine-tuning on Infograph data and evaluating on Sketch data (middle panel of the last row). Using this setup, we proceed to evaluating the impact of intervening on two different aspects of the ImageNet pre-training distribution that have been found to be important in earlier experiments with iWildCam-WILDS: label granularity and data quantity.

\subsection{Label Granularity}
Our setup follows a similar procedure as described previously in Section \ref{sec:label_granularity}. In the left plot of Figure \ref{fig:domainnet_intervention}, when we collapse the label space to 232 superclasses (depth 7), or 17 superclasses (depth 4), the effective robustness obtained from ImageNet pre-training is reduced to about the same level as training from scratch. This suggests that the Infograph-Sketch distribution shift is much more sensitive to label granularity, compared to the iWildCam-WILDS shift.

\subsection{Data Quantity}
We repeat the experiment with pre-training data quantity described in Section \ref{sec:data_quantity}. As seen from the right plot of Figure \ref{fig:domainnet_intervention}, using 25K, 50K, 100K and 150K subsets of ImageNet for pre-training results in a similar level of effective robustness after fine-tuning, which is a lot less than using the full ImageNet dataset, but still an improvement compared to training from scratch. As a sanity check, we find that using only 5K ImageNet images for pre-training leads to no robustness gain compared to training from scratch on Infograph dataset itself.

\end{document}